\documentclass[11pt]{article}

\usepackage[preprint]{acl}

\usepackage{times}
\usepackage{latexsym}

\usepackage[T1]{fontenc}

\usepackage[utf8]{inputenc}

\usepackage{microtype}

\usepackage{inconsolata}

\usepackage{graphicx}

\usepackage[most]{tcolorbox}
\usepackage{amsmath}
\usepackage{amssymb} 
\usepackage{pgfplots}
\pgfplotsset{compat=1.18}
\usepackage{subcaption}  
\usepackage[table]{xcolor}

\usepackage{booktabs}
\usepackage{multirow}

\definecolor{fakegreen}{RGB}{230, 249, 239}  
\usepackage{newfloat}
\usepackage{listings}
\DeclareCaptionStyle{ruled}{labelfont=normalfont,labelsep=colon,strut=off} 

\title{Heterogeneous Strategy Learning in Large Language Model Reasoning: State Tracking and Local Repair}

\author{
    Sunbowen Li\textsuperscript{\rm 1},
    Dandan Song\textsuperscript{\rm 1}\thanks{Corresponding author},
    Shengkun Ma\textsuperscript{\rm 1},
    Xuanning Cao\textsuperscript{\rm 2}  \\ \\
    \textsuperscript{\rm 1}Beijing Institute of Technology,
    \textsuperscript{\rm 2}University of Science and Technology of China
}

\begin{document}
\maketitle
\begin{abstract}
When reasoning traces from different teachers are pooled for supervised fine-tuning (SFT), does a model learn multiple strategies for the same problem, and how faithfully does it execute them? We model each trace as a sample from a latent mixture of strategies and construct graph-search problems with paired breadth-first and depth-first solutions. Under a matched token budget, a 4B model trained without strategy labels generates valid solutions of both strategies on 94.5\% of problems. Across training checkpoints, the mixed-trained model exhibits no mid-search strategy switching before the first error; however, models trained on either mixed or single-strategy data still falter at state tracking, frequently misidentifying whether a state has already been visited. Supplying only the correct event type at this error raises the full-trajectory success rate by 28.00 percentage points, even though unassisted continuations already reach a valid final plan in most cases. Extending local repair to failed natural-language trajectories on GPQA-Diamond and text-only HLE improves final-answer accuracy by 3.79 percentage points across 2,478 error sites; when limited to one repair per trajectory, probe-based selection outperforms random selection, while fixing the earliest error achieves the highest observed accuracy. Overall, our findings show that standard SFT can internalize distinct reasoning strategies without explicit labels, while residual failures stem mainly from local execution errors that can be mitigated through targeted step-level interventions. Our experimental environment is open-sourced at \url{https://github.com/Bowen1911/Symbolic_graph_env}.
\end{abstract}


\section{Introduction}\label{sec:introduction}
Post-training large language models (LLMs) on synthetic reasoning data
commonly pools trajectories from diverse teachers or search algorithms that
solve the same problem via distinct strategies. Although recent methods use
explicit mode labels, prefixes, or specialized objectives to mitigate
interference and preserve diversity
\citep{wu2026mode,jia2026training,fan2026learning}, large-scale pipelines
still routinely train on heterogeneous corpora directly. This raises a
central question: \textbf{can standard autoregressive supervised fine-tuning
(SFT) learn distinct reasoning strategies for the same problem without
explicit mode annotations, and what drives its remaining failures?}

\begin{figure}[t]
    \centering
    \includegraphics[width=\columnwidth]{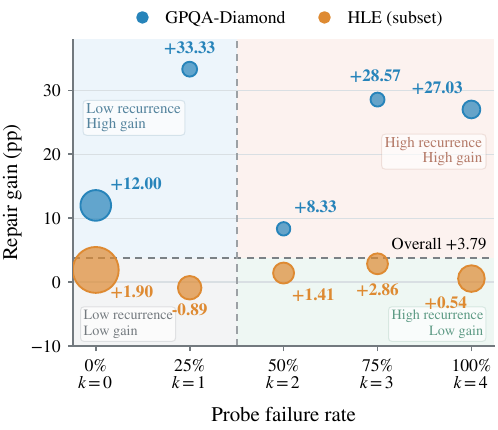}
    \caption{Local error recurrence and repair gain on GPQA-Diamond and a text-only subset of HLE. Here, $k$ denotes the number of failed attempts out of four standalone probe trials; bubble size reflects sample size.}
    \label{fig:recurrence-repair-quadrants}
\end{figure}

We model heterogeneous supervision as a latent strategy mixture where the
expected SFT next-token distribution weights each strategy by its prefix
posterior: shared prefixes preserve multi-strategy branching, whereas
strategy-identifying steps concentrate the posterior onto a single mode. To
test this in finite models, we build a verifiable graph-search environment
pairing each problem with breadth-first (BFS) and depth-first (DFS)
trajectories. Under a matched token budget, unlabeled mixed training enables
Qwen3-4B \citep{yang2025qwen3technicalreport} to reach 99.2\% accuracy and
generate valid BFS and DFS solutions on 94.5\% of problems, matching
explicitly labeled supervision. While multi-strategy competence scales
reliably across 0.6B--14B models and 10K--40K trajectories, smaller models
and earlier checkpoints still leave residual errors.

Do these residual errors stem from mid-trajectory strategy confusion or local
execution slips? Consistent with prefix posterior concentration, executable
replay reveals no strategy switching before the first error; instead,
failures are dominated by state-tracking errors that confuse new and visited
states (e.g., 90.1\% of first errors at the earliest 4B checkpoint), a
bottleneck shared by single-strategy models. Guided by a sensitivity-weighted
bound relating local policy error to continuation value, we test targeted
interventions at these errors: supplying only the correct event type raises
full-trajectory success by 28.00 percentage points (pp), nearly matching
full-step repair, while even unassisted continuations from the
error prefix reach a valid final plan in 87.13\% of cases.

We finally extend local repair to natural-language reasoning on GPQA-Diamond
\citep{rein2024gpqa} and Humanity's Last Exam (HLE) \citep{phan2025humanitys}.
Across all 2,478 localized errors, step-level repairs raise final-answer
accuracy by 3.79 pp overall (+14.78 pp on GPQA-Diamond and +1.64 pp on HLE).
Standalone probing shows that
most errors are low-recurrence execution slips rather than persistent
bottlenecks, yet repairing either type yields substantial gains on
GPQA-Diamond while remaining limited by downstream errors on HLE
(Figure~\ref{fig:recurrence-repair-quadrants}). Under a one-repair-per-trajectory
budget, probe-guided selection outperforms random selection (+3.91 pp), whereas
simply repairing the earliest error achieves the highest accuracy (18.24\%)
without probing overhead. Our main contributions are threefold:
\begin{itemize}
    \item We formulate heterogeneous SFT as a latent strategy mixture and verify
    in symbolic graph search that unlabeled training learns both BFS and DFS
    for the same problem.
    \item Through executable replay and a sensitivity-weighted bound, we show
    that residual errors stem from state tracking rather than strategy
    switching, and are largely resolved by minimal event-type hints (+28.00 pp).
    \item On GPQA-Diamond and HLE, we show that step-level repairs improve
    final accuracy (+3.79 pp), with earliest-error repair outperforming both
    random and probe-guided selection without probing overhead.
\end{itemize}

\section{Related Work}

\subsection{Scaling Laws}
Scaling laws characterize how LLM performance varies with model size, training data, and compute \citep{kaplan2020scaling,hoffmann2022training}.
Recent studies extend scaling analysis to data composition and mixture optimization \citep{ye2025data,shukor2025scaling}.

\subsection{Symbolic Reasoning and Search}
\paragraph{Search and Verification.}
Controlled benchmarks study reasoning with explicit rules and state transitions:
PlanBench evaluates planning and reasoning about change \citep{valmeekam2023planbench},
while ProofWriter generates implications and verifiable proofs from facts and
rules \citep{tafjord2021proofwriter}. NLGraph evaluates graph problems in natural
language \citep{wang2023can}, and NLGift examines generalization beyond pattern
memorization \citep{zhang2024graph}.
Stream of Search learns explicit search and backtracking from algorithm-generated
trajectories \citep{gandhi2024stream}, while Searchformer learns from serialized
A* search dynamics \citep{lehnert2024searchformer}. Recent work
analyzes conditions under which Transformers can represent and learn DFS
\citep{ma2026graphsearch}. \citet{li2026trajectories} recombine partial solutions
on a reasoning graph for verifier-guided inference.

\paragraph{State Tracking.}
Controlled permutation-composition tasks reveal distinct learned state-tracking
mechanisms with different convergence and generalization behavior
\citep{li2025state}. StateBench provides a hierarchy of state-tracking
tasks for evaluating architectural capabilities and length generalization
\citep{petty2026statebench}.

\paragraph{Reasoning Modes.}
Test-time scaling studies highlight the importance of preserving diverse reasoning modes.
ModC \citep{wu2026mode} improves multi-sample reasoning through conditioned training and inference.
SSFT uses a set-based global objective and
bipartite matching to preserve diverse reasoning trajectories and produce
emergent global forking tokens \citep{jia2026training}.
Prefix-conditioned SFT \citep{fan2026learning} mitigates optimization
interference through response-specific prefixes.

\subsection{Importance of Critical Reasoning Steps}
Faith and Fate analyzes compositional reasoning through computation graphs and
step-level errors \citep{dziri2023faith}. The Complexity Ceiling Benchmark
studies sequential reasoning under depth scaling and distinguishes final-answer
correctness from intermediate-trace validity \citep{chapra2026ceiling}.
Prior work uses process-level supervision to evaluate intermediate reasoning steps \citep{lightman2024lets}. Math-Shepherd \citep{wang2024math} uses the outcomes of sampled completions to construct step-level supervision.
Recent studies further reveal that a small fraction of high-entropy tokens can play a disproportionate role in reasoning-oriented reinforcement learning \citep{wang2025beyond}.
Critical Token Fine-Tuning selects consequential tokens for targeted
fine-tuning \citep{ruan2025enhancing}. GPO
identifies critical steps through advantage estimates, samples new
rollouts from these positions, and prioritizes them during optimization
\citep{yu2025gpo}.
Other studies investigate local learnability in multi-teacher distillation \citep{just2025distilling} and the role of reusable reasoning modules in compositional generalization \citep{kong2026reasoning}.

\section{Intrinsic Heterogeneous Learning Formulation}\label{intrinsic-heterogeneous-learning}

To motivate our empirical study, we formulate heterogeneous reasoning as a latent mixture distribution and clarify why standard autoregressive factorization is representationally compatible with unlabeled multi-mode supervision.

\subsection{Multi-Mode Target Distribution}
\label{sec:target-distribution}

For a problem $x\sim p_X$, draw a latent mode $Z$ once per trajectory.
Mode $z$ has probability $\pi_z(x)$ and assigns probability $p_z(y\mid x)$
to a complete trajectory $y=(y_1,\ldots,y_T)$ ending in EOS. Pooling
trajectories without mode labels gives the target distribution
$p^*(y\mid x)=\sum_{z=1}^{K}\pi_z(x)\,p_z(y\mid x)$.

If the model can represent this mixture, it minimizes the expected
SFT loss by matching $p^*(\cdot\mid x)$ for $p_X$-almost every $x$
(Appendix~\ref{app:population-sft}). Whether training on a finite
sample recovers both search strategies is an empirical question.

\subsection{Prefix-Conditioned Mode Posteriors}
\label{sec:implicit-mode-inference}

Let $h_t=(x,y_{<t})$ denote the generation history at step $t$.
For any history with positive probability under $p^*$, the target
next-token distribution is the posterior-weighted mixture
$p^*(y_t\mid h_t)=\sum_{z=1}^{K}p^*(z\mid h_t)p_z(y_t\mid h_t)$,
with Bayes posterior weights:
\begin{equation}
    p^*(z\mid h_t)
    =\frac{\pi_z(x)p_z(y_{<t}\mid x)}
    {\sum_{j=1}^{K}\pi_j(x)p_j(y_{<t}\mid x)},
    \label{eq:mode-posterior}
\end{equation}
where $p_z(y_{<t}\mid x)$ is the prefix likelihood under mode $z$.
For a mode $z^*$ with positive weighted prefix likelihood, its posterior
is one when all other weighted likelihoods are zero, and close to one
when $\sum_{z\ne z^*}\pi_z(x)p_z(y_{<t}\mid x)$ is small relative to
$\pi_{z^*}(x)p_{z^*}(y_{<t}\mid x)$. For deterministic BFS and DFS, a prefix past the first distinguishing
expansion rules out the alternative strategy, concentrating the target
posterior on a single mode; Section~\ref{sec:heterogeneous-supervision}
tests whether finite models learn this behavior. By contrast, training with
explicit strategy labels minimizes $-\mathbb{E}_{x,z,y}\log p_\theta(y\mid x,z)$,
which does not directly constrain the unlabeled distribution; removing the label
at inference is therefore not in general equivalent to marginalizing over
strategies, $\sum_{z=1}^{K}\pi_z(x)p_\theta(y\mid x,z)$.

\section{Multi-Mode Graph Search Learning}

\subsection{Symbolic Search Environment}
\label{sec:symbolic-environment}

We use a symbolic resource-planning environment as a controlled proxy
for studying how LLMs learn and combine search strategies. It isolates
search decisions and state tracking through programmatically generated
tasks, search traces, and executable correctness checks.

\begin{figure}[t]
    \centering
    \includegraphics[width=0.95\columnwidth]{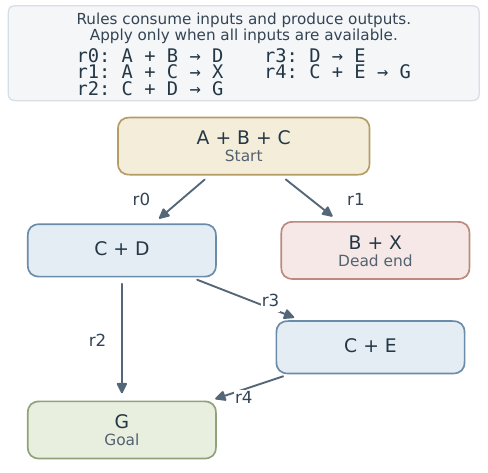}
    \caption{A resource-planning task. Rule \texttt{r1} leads to a dead end.
    Both \texttt{r0,r2} and \texttt{r0,r3,r4} reach the goal.}
    \label{fig:symbolic-demo}
\end{figure}

\paragraph{Task.}
A task specifies an initial resource inventory $\mathbf{s}_0$, a goal
$\mathbf{g}$, and a set of conversion rules $\mathcal{R}$. Each state is a
vector of nonnegative integer resource counts. A rule $r$ consumes
$\mathbf{u}_r$ and produces $\mathbf{v}_r$, updating the state as
$\mathbf{s}_{t+1}=\mathbf{s}_t-\mathbf{u}_r+\mathbf{v}_r$ whenever
$\mathbf{s}_t\succeq\mathbf{u}_r$.
All inequalities are componentwise. The goal is reached when
$\mathbf{s}_t\succeq\mathbf{g}$. Rules do not increase the total resource
count, yielding a finite reachable state space. Figure~\ref{fig:symbolic-demo}
shows an example: each node represents a complete resource inventory, and each
edge applies one rule. Different choices can lead to a solution or a dead end.

\begin{figure}[t]
    \centering
    \includegraphics[width=0.95\columnwidth]{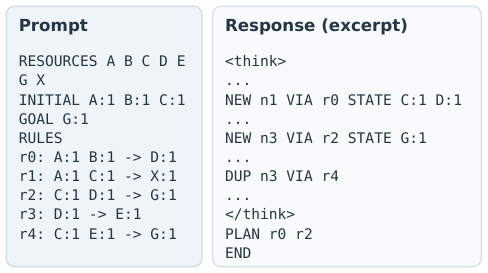}
    \caption{A minimal QA example. Chat role delimiters are omitted;
    \texttt{<think>} is supplied as the assistant prefix.}
    \label{fig:symbolic-qa}
\end{figure}

\paragraph{Search traces.}
The environment supports forward search from $\mathbf{s}_0$ and backward
regression from $\mathbf{g}$, each under BFS or DFS; our experiments pair
forward BFS and DFS traces for the same questions. Inputs list the resource
types, initial inventory, goal, and rules, plus a strategy tag in the labeled
comparison. As shown in Figure~\ref{fig:symbolic-qa}, each expansion selects the
next frontier node via \texttt{EXPAND n STATE} $\mathbf{s}$. If
$\mathbf{s}\succeq\mathbf{g}$, the trace emits \texttt{GOAL n}, outputs an
executable rule sequence (\texttt{PLAN}), and terminates (\texttt{END}).
Otherwise, it applies all valid rules in order: \texttt{NEW n' VIA r STATE}
$\mathbf{s}'$ registers a previously unseen state as a new frontier node
\texttt{n'}, whereas \texttt{DUP n' VIA r} references the first existing node
\texttt{n'} with that state without creating a duplicate node, followed by
\texttt{CLOSE n} to finish the expansion. Generation parameters control resource
counts, rule counts, and dependency structure;
Appendix~\ref{app:symbolic-search-language} and
Table~\ref{tab:search-event-semantics} detail the full event syntax and
semantics.

\paragraph{Verification.}
The verifier replays the search events---checking expansion order, complete rule
enumeration, and \texttt{NEW}/\texttt{DUP} state labels---and independently
executes the final plan. Joint accuracy requires both a legal search trace and
an executable, goal-reaching final plan. Plan-only accuracy checks the latter
condition.

\subsection{Learning from Heterogeneous Supervision}
\label{sec:heterogeneous-supervision}

We compare BFS-only, DFS-only, Mixed, and Mixed+label supervision,
initializing each run from Qwen3-4B-Base. All use the same 20K training
problems and approximately 600M cumulative full-sequence training tokens.
Mixed contains both trajectories per problem; Mixed+label adds an
explicit ``\texttt{STRATEGY BFS/DFS}'' input condition.

We evaluate joint accuracy on 1,000 problems (500 Easy, 500 Medium) and dual-mode coverage
(Cov.@8) on 200 problems. Coverage requires a jointly correct,
exclusively BFS trace and a jointly correct, exclusively DFS trace for the same problem.
The three unlabeled models use eight unconditioned samples per problem;
Mixed+label uses four samples under each strategy condition. Its
accuracy evaluation assigns one label per problem, balanced within
difficulty tiers. Checkpoints are evaluated every 100M training tokens;
T90(obs.) denotes the cumulative training tokens at the first observed
checkpoint reaching 90\% joint accuracy. Appendix~\ref{app:supervision-comparison} gives
training and evaluation details.

\begin{table}[t]
\centering
\setlength{\tabcolsep}{3pt}\scalebox{1}{
\begin{tabular}{@{}lrrr@{}}
\toprule
Training data & \shortstack{Acc.(\%)} & \shortstack{T90(obs.)}
& \shortstack{Cov.@8} \\
\midrule
BFS-only    & 98.8 & 400M & 0/200 \\
DFS-only    & 97.7 & 300M & 0/200 \\
Mixed       & 99.2 & 300M & 189/200 \\
Mixed+label & 98.5 & 300M & 189/200 \\
\bottomrule
\end{tabular}}
\caption{Qwen3-4B at approximately 600M training tokens.
Joint accuracy uses 1,000 validation problems. For Cov.@8, Mixed+label
uses four samples per label, and each other run uses eight unconditioned samples.}
\label{tab:supervision-comparison}
\end{table}

Mixed supervision reaches 99.2\% joint accuracy without accuracy degradation
relative to single-strategy or explicitly labeled training
(Table~\ref{tab:supervision-comparison}). In the Medium subset,
BFS-only, Mixed, and DFS-only reach 71.6\%, 79.8\%, and 86.6\% at 200M tokens
and 83.4\%, 92.2\%, and 93.4\% at 300M tokens (Table~\ref{tab:w1_medium} in
Appendix~\ref{app:supervision-comparison}). Both Mixed and Mixed+label cover
189 of 200 problems (94.5\%).

Interestingly, when we strip the strategy label from the label-trained
Mixed+label model at inference time, it still reaches 97.9\% joint accuracy
(vs.\ 98.5\% with labels), showing that task competence does not depend on the
label token itself. However, 960 of its 979 jointly correct unlabeled responses
follow BFS exclusively (with 19 compatible with both and zero exclusively DFS).
Consistent with Section~\ref{sec:implicit-mode-inference}, omitting inference
labels from a conditioned model does not recover the balanced training mixture;
we note this unconditioned BFS preference as an intriguing side observation
without further verifying its cause (Appendix~\ref{app:supervision-comparison}).

\subsection{Scaling Behavior}
\label{sec:symbolic-training}

\begin{figure}[t]
    \centering
    \includegraphics[width=1\columnwidth]{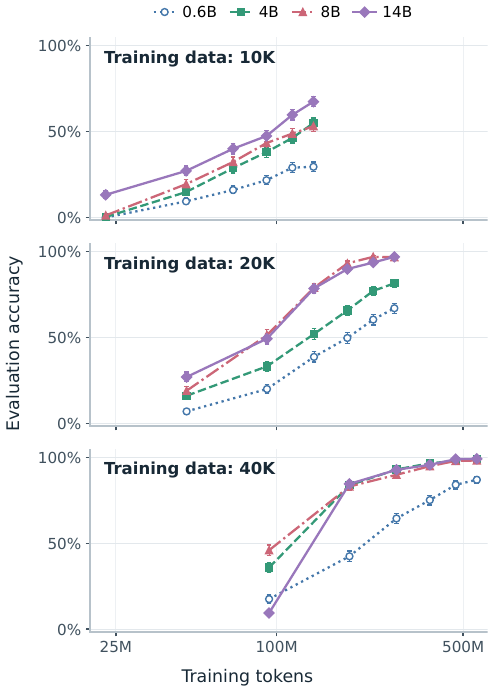}
    \caption{Learning curves for 10K, 20K, and 40K total training
    trajectories, with BFS and DFS mixed at a 1:1 ratio by sequence count.
    Each point represents a training epoch checkpoint.}
    \label{fig:symbolic-scaling-curves}
\end{figure}

We fine-tune Qwen3 Base models of 0.6B, 4B, 8B, and 14B with
unlabeled mixed supervision on 10K, 20K, and 40K trajectories
(5K, 10K, and 20K distinct problems). Each of the 12 runs trains for
six epochs on 8 H20 GPUs, yielding 72 checkpoints evaluated on the same
1,000 validation problems (Figure~\ref{fig:symbolic-scaling-curves}). Increasing the dataset size also increases
cumulative tokens and optimizer updates. Appendix~\ref{app:scaling-experiments}
details the training setup.

On 40K trajectories, the 0.6B model reaches 84.1\% joint accuracy at
epoch five, while the larger models reach at least 98.0\%
(Table~\ref{tab:bfs_dfs_confusion} in Appendix~\ref{app:mode-confusion}). Larger models yield roughly balanced
BFS and DFS shares at the larger budgets among successful traces with
an identifiable strategy (Figure~\ref{fig:marginal-mode-ratio} in Appendix~\ref{app:mode-confusion}).

\section{State Tracking and Local Repair}
\label{sec:local-bottlenecks}

Dual-mode coverage shows that a single model can learn both strategies without
labels. To determine whether remaining failures stem from mid-trajectory
strategy confusion or local execution errors, we replay individual trajectories
to locate their first errors and test targeted local interventions.

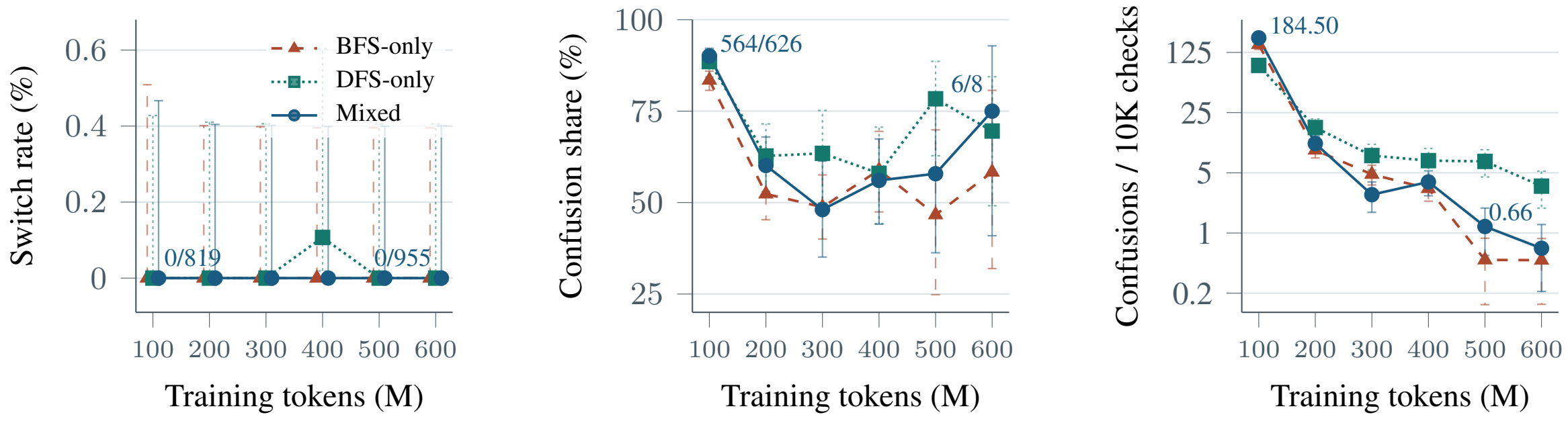
\begin{figure*}[t]
    \centering
\definecolor{navy}{HTML}{195B83}
\definecolor{brick}{HTML}{AB482D}
\definecolor{teal}{HTML}{15786C}
\definecolor{slate}{HTML}{495962}
\definecolor{gridgray}{HTML}{DCE3E7}
\pgfplotsset{
    paper axis/.style={
        tick align=outside,
        axis x line*=bottom, axis y line*=left,
        axis line style={slate},
        tick style={slate},
        tick label style={font=\footnotesize,text=slate},
        label style={font=\footnotesize},
        title style={font=\footnotesize\bfseries},
        ymajorgrids, major grid style={gridgray},
        scaled ticks=false,
        clip=false
    }
}
    \captionsetup[subfigure]{font=small,justification=centering}
    \begin{subfigure}[t]{0.32\textwidth}
        \centering
        \begin{tikzpicture}
\begin{axis}[paper axis,width=\dimexpr\linewidth-12pt\relax,height=127pt,xticklabel style={font=\scriptsize},xmin=70,xmax=630,ymin=-.09,ymax=.68,ytick={0,.2,.4,.6},xtick={100,200,...,600}, xlabel={Training tokens (M)}, ylabel={Switch rate (\%)},
legend style={at={(0.98,0.98)},anchor=north east,legend columns=1,font=\scriptsize,draw=none,fill=white,fill opacity=0.75,text opacity=1,inner sep=1pt,row sep=-1pt,column sep=3pt,cells={anchor=west}},
legend image post style={scale=0.75}]
\addplot[brick,mark options={solid,fill=brick},dashed,mark=triangle*,mark size=1.8pt,line width=.72pt,error bars/.cd,y dir=both,y explicit,error bar style={line width=.42pt,opacity=.58,brick},error mark options={line width=.42pt,mark size=1.25pt,rotate=90}] coordinates {(90,0.000000) += (0,0.508909) -= (0,0.000000)
(190,0.000000) += (0,0.401054) -= (0,0.000000)
(290,0.000000) += (0,0.398557) -= (0,0.000000)
(390,0.000000) += (0,0.395683) -= (0,0.000000)
(490,0.000000) += (0,0.395683) -= (0,0.000000)
(590,0.000000) += (0,0.395276) -= (0,0.000000)};
\addlegendentry{BFS-only}
\addplot[teal,mark options={solid,fill=teal},densely dotted,mark=square*,mark size=1.8pt,line width=.72pt,error bars/.cd,y dir=both,y explicit,error bar style={line width=.42pt,opacity=.58,teal},error mark options={line width=.42pt,mark size=1.25pt,rotate=90}] coordinates {(100,0.000000) += (0,0.427855) -= (0,0.000000)
(200,0.000000) += (0,0.410482) -= (0,0.000000)
(300,0.000000) += (0,0.406142) -= (0,0.000000)
(400,0.106952) += (0,0.496365) -= (0,0.088070)
(500,0.000000) += (0,0.405285) -= (0,0.000000)
(600,0.000000) += (0,0.405713) -= (0,0.000000)};
\addlegendentry{DFS-only}
\addplot[navy,mark options={solid,fill=navy},solid,mark=*,mark size=1.8pt,line width=.72pt,error bars/.cd,y dir=both,y explicit,error bar style={line width=.42pt,opacity=.58,navy},error mark options={line width=.42pt,mark size=1.25pt,rotate=90}] coordinates {(110,0.000000) += (0,0.466853) -= (0,0.000000)
(210,0.000000) += (0,0.404432) -= (0,0.000000)
(310,0.000000) += (0,0.401893) -= (0,0.000000)
(410,0.000000) += (0,0.398971) -= (0,0.000000)
(510,0.000000) += (0,0.399802) -= (0,0.000000)
(610,0.000000) += (0,0.400635) -= (0,0.000000)};
\addlegendentry{Mixed}

\node[font=\scriptsize,text=navy,anchor=south west,inner sep=1pt] at (axis cs:113,0.02) {0/819};
\node[font=\scriptsize,text=navy,anchor=south east,inner sep=1pt] at (axis cs:600,0.02) {0/955};
\end{axis}
\end{tikzpicture}
        \caption{Strategy consistency}
        \label{fig:matched-switch}
    \end{subfigure}\hfill    \begin{subfigure}[t]{0.32\textwidth}
        \centering
        \begin{tikzpicture}
\begin{axis}[paper axis,width=\dimexpr\linewidth-12pt\relax,height=127pt,xticklabel style={font=\scriptsize},xmin=70,xmax=630,ymin=20,ymax=100,ytick={25,50,75,100},xtick={100,200,...,600}, xlabel={Training tokens (M)}, ylabel={Confusion share (\%)}]
\addplot[brick,mark options={solid,fill=brick},dashed,mark=triangle*,mark size=1.8pt,line width=.72pt,error bars/.cd,y dir=both,y explicit,error bar style={line width=.42pt,opacity=.58,brick},error mark options={line width=.42pt,mark size=1.25pt,rotate=90}] coordinates {(100,83.486239) += (0,2.465603) -= (0,2.801099)
(200,52.356021) += (0,6.966556) -= (0,7.059458)
(300,48.760331) += (0,8.806322) -= (0,8.730031)
(400,58.904110) += (0,10.564618) -= (0,11.454886)
(500,46.666667) += (0,23.216353) -= (0,21.857131)
(600,58.333333) += (0,22.340635) -= (0,26.382202)};
\addplot[teal,mark options={solid,fill=teal},densely dotted,mark=square*,mark size=1.8pt,line width=.72pt,error bars/.cd,y dir=both,y explicit,error bar style={line width=.42pt,opacity=.58,teal},error mark options={line width=.42pt,mark size=1.25pt,rotate=90}] coordinates {(100,88.541667) += (0,3.182071) -= (0,4.196705)
(200,62.745098) += (0,8.759926) -= (0,9.685079)
(300,63.461538) += (0,11.737762) -= (0,13.589860)
(400,58.000000) += (0,12.624997) -= (0,13.766558)
(500,78.378378) += (0,10.235110) -= (0,15.573527)
(600,69.565217) += (0,14.830758) -= (0,20.430974)};
\addplot[navy,mark options={solid,fill=navy},solid,mark=*,mark size=1.8pt,line width=.72pt,error bars/.cd,y dir=both,y explicit,error bar style={line width=.42pt,opacity=.58,navy},error mark options={line width=.42pt,mark size=1.25pt,rotate=90}] coordinates {(100,90.095847) += (0,2.101120) -= (0,2.590217)
(200,60.144928) += (0,7.787198) -= (0,8.336704)
(300,48.076923) += (0,13.237394) -= (0,12.972809)
(400,56.060606) += (0,11.311273) -= (0,11.977971)
(500,57.894737) += (0,18.963372) -= (0,21.618833)
(600,75.000000) += (0,17.852079) -= (0,34.072457)};

\node[font=\scriptsize,text=navy,anchor=south west,inner sep=1pt] at (axis cs:112,90) {564/626};
\node[font=\scriptsize,text=navy,anchor=south east,inner sep=1pt] at (axis cs:592,79) {6/8};
\end{axis}
\end{tikzpicture}
        \caption{First-error composition}
        \label{fig:matched-share}
    \end{subfigure}\hfill    \begin{subfigure}[t]{0.32\textwidth}
        \centering
        \begin{tikzpicture}
\begin{axis}[paper axis,width=\dimexpr\linewidth-12pt\relax,height=127pt,xticklabel style={font=\scriptsize},xmin=70,xmax=630,ymode=log,log basis y=10,ymin=.12,ymax=300,ytick={.2,1,5,25,125},yticklabels={0.2,1,5,25,125},xtick={100,200,...,600}, xlabel={Training tokens (M)}, ylabel={Confusions / 10K checks}]
\addplot[brick,mark options={solid,fill=brick},dashed,mark=triangle*,mark size=1.8pt,line width=.72pt,error bars/.cd,y dir=both,y explicit,error bar style={line width=.42pt,opacity=.58,brick},error mark options={line width=.42pt,mark size=1.25pt,rotate=90}] coordinates {(100,154.694254) += (0,23.342519) -= (0,19.550482)
(200,9.185182) += (0,1.888836) -= (0,1.750340)
(300,4.801159) += (0,1.325842) -= (0,1.192399)
(400,3.282643) += (0,1.001654) -= (0,0.939645)
(500,0.487262) += (0,0.382628) -= (0,0.340047)
(600,0.483589) += (0,0.379528) -= (0,0.334905)};
\addplot[teal,mark options={solid,fill=teal},densely dotted,mark=square*,mark size=1.8pt,line width=.72pt,error bars/.cd,y dir=both,y explicit,error bar style={line width=.42pt,opacity=.58,teal},error mark options={line width=.42pt,mark size=1.25pt,rotate=90}] coordinates {(100,88.348405) += (0,11.256444) -= (0,10.660162)
(200,16.781163) += (0,4.321679) -= (0,3.880934)
(300,7.931167) += (0,2.784775) -= (0,2.559252)
(400,6.926531) += (0,2.649046) -= (0,2.336066)
(500,6.790774) += (0,2.471972) -= (0,2.320282)
(600,3.501171) += (0,1.723492) -= (0,1.557546)};
\addplot[navy,mark options={solid,fill=navy},solid,mark=*,mark size=1.8pt,line width=.72pt,error bars/.cd,y dir=both,y explicit,error bar style={line width=.42pt,opacity=.58,navy},error mark options={line width=.42pt,mark size=1.25pt,rotate=90}] coordinates {(100,184.500638) += (0,24.445392) -= (0,22.466363)
(200,10.936594) += (0,2.491493) -= (0,2.233582)
(300,2.788156) += (0,1.121502) -= (0,1.048651)
(400,3.916421) += (0,1.353817) -= (0,1.209663)
(500,1.189305) += (0,0.759228) -= (0,0.639869)
(600,0.664621) += (0,0.590074) -= (0,0.455156)};

\node[font=\scriptsize,text=navy,anchor=south west,inner sep=1pt] at (axis cs:112,184.5) {184.50};
\node[font=\scriptsize,text=navy,anchor=south east,inner sep=1pt] at (axis cs:592,1.25) {0.66};
\end{axis}
\end{tikzpicture}
        \caption{State-tracking error rate}
        \label{fig:matched-risk}
    \end{subfigure}

    \caption{Matched-budget replay on 1,000 validation problems per checkpoint
    for Qwen3-4B. (\subref{fig:matched-switch}) Within-trace strategy switch rate
    before the first error, conditioned on at least two strategy-identifying
    expansions (horizontally offset in \subref{fig:matched-switch} for visibility).
    (\subref{fig:matched-share}) Share of all first-error trajectories caused by
    strict \texttt{NEW}/\texttt{DUP} confusions. (\subref{fig:matched-risk}) Number of strict confusions per
    10K expected \texttt{NEW}/\texttt{DUP} steps across valid prefixes (log
    scale). Whiskers denote pointwise 95\%
    confidence intervals.}
    \label{fig:matched-replay}
\end{figure*}

\subsection{Locating First Errors}
\label{sec:locating-first-errors}

We replay 24K responses from the four model sizes at epochs 1--6 on the 1,000
validation problems, normalizing event-specific error rates by valid-prefix
opportunities and error shares by total localized first errors through
completion or the first error (Appendix~\ref{app:w3_methods}).

\paragraph{Errors in state tracking.}
Across 24K responses, models rarely misselect the frontier node
(\texttt{EXPAND} error rate 0.0099\%; Table~\ref{tab:first_error_distribution}).
Instead, expected \texttt{NEW}/\texttt{DUP} steps (57.05\% of valid-prefix
opportunities) account for 4,904 of 5,318 first errors (92.22\%). In the strict
subset (3,602/5,318, 67.73\%; 20.67 per 10K steps), the model applies the
expected rule and state update but reverses whether the resulting state is
unseen (\texttt{NEW}) or already visited (\texttt{DUP}). Thus, search failures
stem primarily from local state tracking rather than frontier scheduling or rule
application.

\begin{table}[t]
\centering
\setlength{\tabcolsep}{3.5pt}\scalebox{1}{
\begin{tabular}{@{}lrrr@{}}
\toprule
Event & Opportunities & First errors & Rate (\%) \\
\midrule
\texttt{NEW}    & 928,999 & 2,761 & 0.2972 \\
\texttt{DUP}    & 813,457 & 2,143 & 0.2634 \\
\texttt{PLAN}   & 18,900  & 218   & 1.1534 \\
\texttt{EXPAND} & 639,587 & 63    & 0.0099 \\
\bottomrule
\end{tabular}}
\caption{First errors by expected event type across 24K responses.
The Opportunities column counts events with a valid preceding prefix,
through completion or including the first error. Errors in other fields are very rare and are therefore not listed.}
\label{tab:first_error_distribution}
\end{table}

\paragraph{Matched-budget replay.}
Replaying 18K responses from the 4B BFS-only, DFS-only, and Mixed runs across
the 100M--600M token checkpoints (Figure~\ref{fig:matched-replay};
Tables~\ref{tab:matched-replay-counts}--\ref{tab:matched-switch} in
Appendix~\ref{app:matched-replay}), Mixed exhibits zero strategy switches before
the first error at every checkpoint (0/819 eligible traces at 100M and 0/955 at
600M; Figure~\ref{fig:matched-switch}), and only one switch at 400M and 500M
when replaying past it. Consistent with prefix posterior concentration
(Section~\ref{sec:implicit-mode-inference}), once the initial expansion commits
to BFS or DFS, the model executes that strategy coherently. First errors in
Mixed are likewise dominated by strict \texttt{NEW}/\texttt{DUP} confusions
tracking the single-strategy runs (90.1\% at 100M and 75.0\% [6/8] at 600M;
Figure~\ref{fig:matched-share}), while confusions per 10K steps fall from 184.50
to 0.66 (Figure~\ref{fig:matched-risk}), confirming that state tracking is a
shared execution bottleneck rather than a byproduct of mixed supervision.

\subsection{Local Error and Downstream Sensitivity}
\label{sec:critical-reasoning-nodes}

An error's consequence depends on the outcome being evaluated and on
how the model continues. We use a performance bound to relate these
quantities and define the contrasts used in our experiments.

\paragraph{Sequential decision formulation.}
Let $h_t=(x,y_{<t})$ be the reasoning history and $a_t$ the next
reasoning action ($a_t=y_t$ at token resolution).
We compare the target policy $p^*$ and student policy $p_\theta$
under the same problem distribution $p_X$ and finite horizon $T$ (padding terminated trajectories with EOS). Let $R(x,y)\in\{0,1\}$ (abbreviated as $R$) be the
binary success indicator---joint trace-and-plan or final-plan accuracy in symbolic search, and final-answer correctness in natural language---with
$J(p)=\mathbb{E}_{x\sim p_X,\,y\sim p(\cdot\mid x)}[R(x,y)]$.
The action value under student continuation is
\begin{equation}
    Q_t^\theta(h_t,a)
    =\Pr\!\left(R=1\mid h_t,\operatorname{do}(a_t=a),p_\theta\right),
    \label{eq:action-value}
\end{equation}
where $\operatorname{do}(a_t=a)$ fixes the current action before resuming $p_\theta$.
For history distribution $d_t^{p^*}$ under $p^*$ and $\Delta J=J(p^*)-J(p_\theta)$,
the performance difference lemma \citep{kakade2002approximately} gives
\begin{equation}
    \begin{aligned}
        \Delta J
        &=\sum_{t=1}^{T}\mathbb{E}_{h_t\sim d_t^{p^*}}\Bigl[
        \sum_a\bigl(p^*(a\mid h_t)-p_\theta(a\mid h_t)\bigr)\\
        &\quad\cdot Q_t^\theta(h_t,a)\Bigr].
    \end{aligned}
    \label{eq:performance-difference}
\end{equation}
With state value
$V_t^\theta(h_t)=\sum_a p_\theta(a\mid h_t)Q_t^\theta(h_t,a)=\Pr(R=1\mid h_t,p_\theta)$
and $V_{T+1}^\theta(h_{T+1})=R(x,y)$, Bellman consistency equates the inner sum
in Equation~\ref{eq:performance-difference} to the expected one-step change in
$V_t^\theta$, which telescopes along $p^*$ trajectories to $J(p^*)-J(p_\theta)$.

\begin{table*}[t]
\centering
\setlength{\tabcolsep}{5pt}\scalebox{0.9}{
\begin{tabular}{@{}lrrrrrrrrr@{}}
\toprule
& & \multicolumn{3}{c}{Joint success (\%)} & \multicolumn{2}{c}{Joint C$-$A} &
\multicolumn{3}{c}{Plan-only accuracy (\%)} \\
\cmidrule(lr){3-5}\cmidrule(lr){6-7}\cmidrule(l){8-10}
Model & Cases & A & B & C & pp [95\% CI] & Adjusted $p$ & A & B & C \\
\midrule
0.6B & 495 & 10.15 & 32.22 & 32.47 & $+22.32$ [$+19.14$, $+25.61$] & $<0.0002$ & 86.36 & 86.72 & 87.17 \\
4B & 98 & 28.06 & 63.27 & 66.33 & $+38.27$ [$+31.12$, $+45.92$] & $<0.0002$ & 88.52 & 89.80 & 87.24 \\
8B & 88 & 24.72 & 65.34 & 70.17 & $+45.45$ [$+37.50$, $+53.69$] & $<0.0002$ & 84.09 & 85.23 & 84.94 \\
14B & 86 & 31.69 & 72.67 & 70.35 & $+38.66$ [$+30.23$, $+47.38$] & $<0.0002$ & 93.02 & 93.90 & 93.90 \\
\midrule
Pooled & 767 & 16.53 & 44.52 & 45.37 & $+28.85$ [$+26.21$, $+31.56$] & $<0.0002$ & 87.13 & 87.74 & 87.68 \\
\bottomrule
\end{tabular}}
\caption{Local assistance at first same-rule \texttt{NEW}/\texttt{DUP} confusions.
A resamples unassisted, B supplies the event type, and C supplies the full event.
Each case contributes four continuations per condition. Intervals cluster by source problem.
Holm correction applies to joint-success comparisons. The pooled plan-only contrasts are
$+0.62$ pp [$-0.55$, $+1.79$] for B$-$A and $+0.55$ pp [$-0.62$, $+1.74$] for C$-$A.}
\label{tab:local-assistance}
\end{table*}

\paragraph{Local error and decision sensitivity.}
Define the local policy error and decision sensitivity as
\begin{equation}
    \begin{aligned}
        e_t(h_t)&=D_{\mathrm{TV}}\!\left(
        p^*(\cdot\mid h_t),p_\theta(\cdot\mid h_t)\right),\\
        S_t^\theta(h_t)&=\max_a Q_t^\theta(h_t,a)
        -\min_a Q_t^\theta(h_t,a),
    \end{aligned}
    \label{eq:error-and-sensitivity}
\end{equation}
where $D_{\mathrm{TV}}(p,q)=\tfrac12\sum_a|p(a)-q(a)|$ is the total variation
distance. Since policy differences sum to zero, centering $Q_t^\theta$ at its
range midpoint bounds the absolute inner sum in
Equation~\ref{eq:performance-difference} by $S_t^\theta(h_t)e_t(h_t)$, yielding
\begin{equation}
    |\Delta J|\leq\sum_{t=1}^{T}
    \mathbb{E}_{h_t\sim d_t^{p^*}}\!\left[
    S_t^\theta(h_t)e_t(h_t)\right].
    \label{eq:sensitivity-tv-bound}
\end{equation}
With $\kappa_t(h_t)=D_{\mathrm{KL}}(p^*(\cdot\mid h_t)\Vert p_\theta(\cdot\mid h_t))$,
Pinsker's inequality $e_t(h_t)\leq\sqrt{\kappa_t(h_t)/2}$ further gives
\begin{equation}
    |\Delta J|\leq\sum_{t=1}^{T}
    \mathbb{E}_{h_t\sim d_t^{p^*}}\!\left[
    S_t^\theta(h_t)\sqrt{\kappa_t(h_t)/2}\right].
    \label{eq:sensitivity-kl-bound}
\end{equation}
The product $C_{\mathrm{pot}}(h_t)=S_t^\theta(h_t)e_t(h_t)$ bounds the absolute
local term in Equation~\ref{eq:performance-difference} and motivates a critical
reasoning node, pairing standalone probes of local error recurrence with
targeted interventions measuring downstream recovery.

\begin{figure}[t]
    \centering
    \includegraphics[width=\columnwidth]{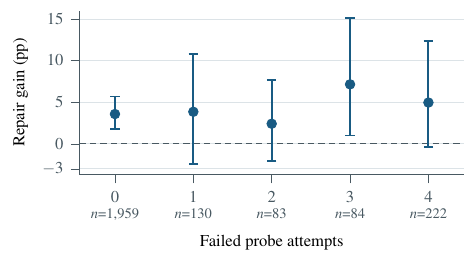}
    \caption{Paired repair gain by failed attempts on the four-sample
    standalone probe. Bars show pointwise 95\% source-problem cluster
    bootstrap intervals. Sample sizes count valid node pairs.}
    \label{fig:probe-repair-gain}
\end{figure}

\paragraph{Local intervention value.}
At event resolution, let $h$ be a valid prefix immediately before an
intervention step, $a_{\mathrm{fix}}$ a correct event, and
$V^\theta(h)=\Pr(R=1\mid h,p_\theta)$ the unassisted continuation success rate.
The single-step local repair value is
\begin{equation}
    \Delta_{\mathrm{repair}}^\theta(h)
    =Q^\theta(h,a_{\mathrm{fix}})-V^\theta(h),
    \label{eq:local-repair-value}
\end{equation}
and for a general assistance procedure $I$, we define
\begin{equation}
    \Delta_I^\theta(h)
    =\Pr(R=1\mid h,I,p_\theta)-V^\theta(h),
    \label{eq:local-assistance-value}
\end{equation}
which reduces to Equation~\ref{eq:local-repair-value} when $I$ fixes a single
action $a_{\mathrm{fix}}$ and extends to natural-language repairs evaluated at
localized error prefixes.

\begin{table*}[t]
\centering
\setlength{\tabcolsep}{5pt}\scalebox{0.88}{
\begin{tabular}{@{}llrrrrrc@{}}
\toprule
Model & Benchmark & \shortstack{Reproducible\\errors (\%)} & \shortstack{Valid A/C\\pairs} & \shortstack{A: Unasst.\\(\%)} & \shortstack{C: Repaired\\(\%)} & \shortstack{C$-$A\\(pp)} & \shortstack{95\% CI\\(pp)} \\
\midrule
Qwen3.8-27B & GPQA-D & 52/295 (17.63) & 41 & 17.07 & 41.46 & +24.39 & [+6.06, +43.90] \\
Qwen3.8-27B & HLE & 324/1,642 (19.73) & 272 & 5.15 & 5.88 & +0.74 & [$-$2.75, +3.90] \\
DeepSeek & GPQA-D & 27/290 (9.31) & 22 & 4.55 & 27.27 & +22.73 & [+3.84, +55.56] \\
DeepSeek & HLE & 94/970 (9.69) & 54 & 3.70 & 7.41 & +3.70 & [+0.00, +9.62] \\
\midrule
Total ($k\geq 2$) & --- & 497/3,197 (15.55) & 389 & 6.17 & 11.05 & +4.88 & [+0.86, +10.56] \\
Total ($k<2$) & --- & 2,700/3,197 (84.45) & 2,089 & 5.84 & 9.43 & +3.59 & [+1.82, +5.66] \\
\bottomrule
\end{tabular}}
\caption{Repair gains among localized errors by probe recurrence ($k\geq 2$ by model and benchmark, plus $k<2$ total; all 2,478 valid pairs average $+3.79$ pp).
    Qwen and DeepSeek denote Qwen3.8-27B and DeepSeek V4.1 Flash. Valid
    pair counts exclude unavailable, answer-revealing, or
    incomplete repairs.}
\label{tab:crn-nl-tasks}
\end{table*}

\begin{table}[t]
\centering
\setlength{\tabcolsep}{3pt}\scalebox{0.83}{
\begin{tabular}{@{}lrrrr@{}}
\toprule
Sample & $n$ & Earliest & Random & Probe \\
\midrule
All with common C & 329 & 18.24 & 11.29 & 15.20 \\
Multiple sites & 301 & 18.60 & 11.01 & 15.28 \\
\midrule
\multicolumn{5}{@{}l}{Probe contrasts (pp), with 95\% CIs} \\
& & \multicolumn{1}{r}{vs.\ earliest} & \multicolumn{2}{c}{vs.\ random} \\
All with common C & &
    \shortstack{$-3.04$\\$[-6.47, +0.28]$} &
    \multicolumn{2}{c}{\shortstack{$+3.91$\\$[+1.30, +6.77]$}} \\
\addlinespace[2pt]
Multiple sites & &
    \shortstack{$-3.32$\\$[-7.14, +0.31]$} &
    \multicolumn{2}{c}{\shortstack{$+4.28$\\$[+1.42, +7.47]$}} \\
\bottomrule
\end{tabular}}
\caption{Final-answer accuracy (\%) and paired contrasts in the
one-repair replay. Intervals cluster by source problem.}
\label{tab:one-repair}
\end{table}

\subsection{Continuation after Local Repairs}
\label{sec:local-assistance}
From the epoch-2 checkpoints trained on 40K traces, we select all 767
model--problem cases whose first error is a same-rule
\texttt{NEW}/\texttt{DUP} confusion. Resuming from the valid prefix before the
error, the original checkpoint generates four continuations per case under
unassisted resampling (A), supplying the correct event type (B), and supplying
the full correct event (C).

Across all cases, the event-type hint raises joint success from 16.53\% to
44.52\% (+28.00 pp), recovering 97\% of the full-event repair gain (45.37\%,
+28.85 pp; Table~\ref{tab:local-assistance}; Appendix~\ref{app:local-assistance}).
Meanwhile, unassisted resampling already achieves 87.13\% plan-only accuracy,
and local assistance changes it by less than 1 pp. Thus, models already possess
the downstream competence to complete the search once disambiguated on binary
state novelty (\texttt{NEW} vs.\ \texttt{DUP}), and because graph search permits
alternative valid paths, local state-tracking errors primarily corrupt
intermediate process validity rather than final-plan reachability.

\subsection{Local Repairs in Natural Language}
\label{sec:crn-natural-language}
To test whether local repairs also rescue final answers when errors
propagate along natural-language derivations, we collect four baseline rollouts
per problem on GPQA-Diamond and a 100-problem text-only HLE subset from
Qwen3.8-27B \citep{qwen2026qwen38} and DeepSeek V4.1 Flash
\citep{deepseekai2026deepseekv41flash}. GPT-6-Astra localizes 3,197 errors from
failed trajectories and proposes minimal repairs (audit details in
Appendix~\ref{app:crn-nl-details}).

\paragraph{Repair gain and probe recurrence.}
An atomic probe reformulates each localized decision as a standalone
short-answer question with the required premises and records the number of
failed attempts $k\in\{0,\ldots,4\}$ across four samples; only 497 of the 3,197
sites (15.55\%) have $k\geq 2$. At each eligible site, we compare unassisted
continuation from the original prefix (A) against continuation with the local
repair appended (C). Across all 2,478 valid pairs (406 GPQA-D, $+14.78$ pp;
2,072 HLE, $+1.64$ pp), local repair improves final-answer accuracy by 3.79 pp
overall, with gains of 4.88 pp for $k\geq 2$ and 3.59 pp for $k<2$
(Table~\ref{tab:crn-nl-tasks}, Figure~\ref{fig:probe-repair-gain};
Appendix~\ref{app:probe-repair}). Gains concentrate on GPQA-Diamond ($+24.39$
and $+22.73$ pp for $k\geq 2$, vs.\ $+0.74$ and $+3.70$ pp on HLE), where high
baseline competence ($>89\%$ vs.\ $25\%\text{--}35\%$) lets models finish the
derivation once the local error is fixed, whereas HLE trajectories often hit
subsequent bottlenecks. Thus, most trajectory errors are
in-context execution slips ($k<2$) rather than persistent atomic blind spots
($k\geq 2$), yet fixing either type at a downstream-sensitive step recovers
correct final answers, making standalone probe recurrence only a weak predictor
of repair gain.

\paragraph{Selecting one repair site.}
When limited to a single repair per failed trajectory, we compare three
selection policies in retrospective replay: repairing the earliest eligible
error, selecting an eligible error uniformly at random, and selecting the error
with the highest probe failure count $k$.

Across 329 trajectories, repairing the earliest error achieves the highest
final-answer accuracy (18.24\%), followed by probe selection (15.20\%) and
random selection (11.29\%; Table~\ref{tab:one-repair}). On the 326 trajectories
with complete A/C outcomes, the earliest and probe policies yield similar net
repair gains (differing by 0.61 pp), but earlier prefixes retain higher baseline
continuation accuracy (10.12\% vs.\ 6.75\%;
Appendix~\ref{app:one-repair-replay}). Consequently, while probe selection
beats random selection (+3.91 pp), error position matters more than standalone
recurrence: repairing the earliest error prevents upstream mistakes from
compounding and achieves the highest accuracy without the 9,820 extra probe calls.

\section{Conclusion}
We studied whether standard autoregressive SFT can learn distinct reasoning
strategies for the same problem without explicit mode labels. Modeling each
trajectory as a draw from a latent strategy mixture explains how the generated
prefix commits the continuation to a single mode, and paired graph-search
experiments confirm that unlabeled training produces both valid BFS and DFS
solutions for the same problem. Trajectory replay further shows that models
trained on mixed or single-strategy data maintain consistent expansion order
yet falter at state tracking; supplying only the correct event type
substantially raises full-trajectory success, even though
unassisted continuations already reach a valid final plan in most cases.
Extending local repair to GPQA-Diamond and HLE improves final-answer
accuracy by 3.79 pp, and under a single-repair
budget, fixing the earliest error outperforms both random and probe-guided
selection without probing overhead. Overall, standard SFT can internalize
distinct reasoning strategies without explicit annotations, while residual
failures stem mainly from local execution errors that can be mitigated through
targeted step-level interventions.

\section*{Limitations}

\paragraph{Scope of the learning comparison.}
Each symbolic problem has paired BFS and DFS traces generated under
deterministic expansion rules. An informative expansion can identify
the strategy exactly. Natural-language solutions can share long
prefixes or legitimately switch strategies at a subgoal. The four-way
supervision comparison uses one model size (4B) and one training seed.
The cross-scale runs use mixed supervision throughout. Our results
therefore establish dual-mode coverage within the tested sampling budget,
while the relative merits of labeled and unlabeled training across seeds,
model sizes, and natural-language strategies remain open.

\paragraph{Observability and choice of outcome.}
A trajectory can reveal a switch before its first error only if it
contains at least two strategy-identifying expansions. Mistakes that
terminate a trace early can remove it from this comparison. The
earliest matched checkpoint is at 100M tokens, and the final Mixed
checkpoint has only eight first errors. This limits both the timing
we can infer for changes in strategy use and the precision of late
error shares. The repair experiment further conditions on a specific
first error. Its gain of 28 pp in joint success primarily reflects a
more faithful search record. Interpreting this gain as an improvement
in final-answer reliability would overstate the observed effect on
plan accuracy.

\paragraph{Natural-language repair and selection.}
Eligibility requires a failed trajectory, a localized error, and a
repair that passes the answer-leakage filter. A standalone probe can
supply premises missing in the original prefix, and a repair can add
information beyond the corrected decision. Localization and grading
also rely on models. In the independent audit, only 6 of 11 repairs
at confirmed Qwen GPQA-Diamond error sites were validated
(Table~\ref{tab:astra-verification} in Appendix~\ref{app:crn-nl-details}). The 2,478 paired sites share 93
source problems. Clustering accounts for that dependence, but a single
continuation per condition leaves each site's repair probability poorly
determined. The policy replay reuses these problems and breaks probe ties
by choosing the earliest site. Even after skipping single-candidate
trajectories, probe selection required 9,820 additional generations.
Testing whether this cost yields an advantage on new problems calls for
a held-out comparison with repeated continuations. Using repair signals
to weight training examples would require a separate training experiment.


\bibliography{custom}

\appendix\label{sec:appendix}

\section{Expected SFT Loss for the Target Mixture}
\label{app:population-sft}
For the trajectory mixture in Section~\ref{sec:target-distribution},
$\pi_z(x)\geq 0$ and $\sum_{z=1}^{K}\pi_z(x)=1$.
The expected SFT loss is the negative
log-likelihood of complete trajectories,
$\mathcal{L}(\theta)=-\mathbb{E}_{x\sim p_X,\,y\sim p^*(\cdot\mid x)}\log p_\theta(y\mid x)$.
Assuming finite conditional entropy, this cross-entropy decomposes into
the target conditional entropy and the expected KL divergence:
\begin{equation}
    \begin{aligned}
        \mathcal{L}(\theta)
        &=H_{p^*}(Y\mid X)\\
        &\quad+\mathbb{E}_{x\sim p_X}D_{\mathrm{KL}}\!\left(
        p^*(\cdot\mid x)\Vert p_\theta(\cdot\mid x)\right).
    \end{aligned}
    \label{eq:risk-decomposition}
\end{equation}
Measuring distribution learning by the excess risk
$E_{\mathrm{dist}}(\theta)=\mathcal{L}(\theta)-H_{p^*}(Y\mid X)$---which
equals the average KL divergence from $p^*$ to $p_\theta$---implies that
when the model can represent $p^*$ and reaches a global minimum of
$\mathcal{L}(\theta)$, we have $E_{\mathrm{dist}}(\theta)=0$ and hence
$p_\theta(\cdot\mid x)=p^*(\cdot\mid x)$ for $p_X$-almost every $x$.

\section{Symbolic Search Language}
\label{app:symbolic-search-language}

\paragraph{Vector fields and direction.}
Vectors contain nonnegative integer resource counts. In the dense
notation below, entries follow the input \texttt{RESOURCES} order and
intermediate zeros must be retained. The saved QA example in
Appendix~\ref{app:qa-example} instead uses named, sparse entries of the
form \texttt{B:1}; unlisted resources have zero counts.
\texttt{STATE} denotes the resources currently held in
forward search, whereas \texttt{NEED} denotes outstanding requirements
in backward search. The forward root is $\mathbf{s}_0$, and the backward
root is $\mathbf{g}$. A trajectory uses only one of these vector keywords.
The keyword identifies search direction; BFS versus DFS is determined
by the actual expansion order, using FIFO versus LIFO scheduling.

\paragraph{Event syntax.}
Let \texttt{n} be a node identifier, \texttt{r} a rule identifier,
and \texttt{V} a placeholder for \texttt{STATE} or \texttt{NEED}.
For $d$ resource types, the event forms are:
\begin{tcolorbox}[breakable,title={Search event syntax},fonttitle=\bfseries\small,
    colback=black!2,colframe=black!50,colbacktitle=black!8,coltitle=black,
    boxrule=0.4pt,arc=1mm,left=6pt,right=6pt,top=4pt,bottom=4pt]
\begin{lstlisting}[basicstyle=\ttfamily\small,columns=fullflexible,keepspaces=true,breaklines=true]
EXPAND n V c1 ... cd
NEW n VIA r V c1 ... cd
DUP n VIA r
CLOSE n
GOAL n
PLAN r1 ... rk
END
\end{lstlisting}
\end{tcolorbox}
Here, \texttt{c1 ... cd} supplies the complete vector, and \texttt{VIA}
specifies the rule applied from the currently expanded node.
Table~\ref{tab:search-event-semantics} summarizes the event meanings.

\begin{table}[t]
\centering
\small
\setlength{\tabcolsep}{4pt}
\begin{tabular}{@{}p{0.18\columnwidth}p{0.78\columnwidth}@{}}
\toprule
Event & Meaning \\
\midrule
\texttt{EXPAND} & Remove node \texttt{n} from the frontier, check its
vector and the termination condition, and increment the check count. \\
\addlinespace
\texttt{NEW} & Apply rule \texttt{r} to the current node and register
a previously unseen vector as node \texttt{n}. \\
\addlinespace
\texttt{DUP} & The result of rule \texttt{r} matches an existing vector.
Reference its first registered node \texttt{n}; do not create a new node. \\
\addlinespace
\texttt{CLOSE} & Finish enumerating all rules for the current
nonterminal node and commit the frontier update, even if no successors
were produced. \\
\addlinespace
\texttt{GOAL} & The current node satisfies the direction-specific
termination condition. No \texttt{NEW}, \texttt{DUP}, or \texttt{CLOSE}
events follow for this node. \\
\addlinespace
\texttt{PLAN} & Output the final rule sequence in forward execution order. \\
\addlinespace
\texttt{END} & Mark the end of the trajectory. \\
\bottomrule
\end{tabular}
\caption{Meanings of symbolic search events. Event lists the output
keyword; Meaning describes its operation. In the arguments,
\texttt{n} identifies a node and \texttt{r} a conversion rule.
Vectors describe held resources in forward search (\texttt{STATE})
or outstanding requirements in backward search (\texttt{NEED}).}
\label{tab:search-event-semantics}
\end{table}

\paragraph{Termination and plan direction.}
Forward search terminates when the current state satisfies
$\mathbf{s}\succeq\mathbf{g}$; backward search terminates when
$\mathbf{s}_0\succeq\mathbf{q}$, where $\mathbf{q}$ is the current
\texttt{NEED} vector. Both directions output a forward-executable
\texttt{PLAN}. In backward search, the rule sequence along the parent
chain from the root to the goal node is reversed to obtain this plan.
For example, a backward parent chain labeled \texttt{r1 r0} produces
\texttt{PLAN r0 r1}.

\paragraph{Rule enumeration and verification.}
No event is emitted for a rule that is inapplicable in forward search,
or whose backward regression exceeds the total requirement bound $m$.
The verifier independently computes all expected \texttt{NEW} and
\texttt{DUP} events. At \texttt{CLOSE}, it checks for omitted rules,
incorrect ordering, and spurious nodes. Thus, the trace need not print
the entire frontier, but a single path cannot substitute for a complete
BFS trace. Under the rule constraints
$\mathbf{u}_r\neq\mathbf{v}_r$ and
$\|\mathbf{v}_r\|_1\leq\|\mathbf{u}_r\|_1$, neither a forward transition
nor a backward regression reproduces the expanded node's own vector;
a \texttt{DUP} therefore references another registered node.

\paragraph{Search budget.}
The check budget is tested before removing a node from the frontier.
If it is exhausted, search stops after the preceding \texttt{CLOSE},
without emitting an incomplete \texttt{EXPAND} event.
Budget-exhausted runs are failures: their valid event prefixes may be
retained for diagnostics, but are excluded from successful QA examples.

\section{Training and Evaluation Details}
\label{app:symbolic-training-details}

\subsection{Matched-Token Supervision Comparison}
\label{app:supervision-comparison}

\paragraph{Data and token accounting.}
The four configurations in Section~\ref{sec:heterogeneous-supervision}
start independently from Qwen3-4B-Base and share 20K distinct training
problems, half easy and half medium. BFS-only and DFS-only repeat only
the corresponding strategy's trace for these problems. Mixed and
Mixed+label use the same paired trajectories, with the latter adding
\texttt{STRATEGY BFS} or \texttt{STRATEGY DFS} as the first user-input
line. All groups use the same problem order within a traversal; mixed
groups swap each problem's strategy between adjacent traversals.

The target budget is 600M native full-sequence tokens, including the
prompt, thinking trace, \texttt{PLAN}, and native template. Complete
sample groups are retained, allowing small overshoots of the target.
We track actual cumulative tokens, sample counts, problem
coverage, and supervised tokens separately. Trace lengths and label
overhead change the number of consumed samples, problem repetitions,
and stopping points, so matching tokens does not match per-problem pass counts.
Sequence-length distributions also affect attention computation.

\paragraph{Optimization and checkpoints.}
Each run uses training BF16, Adam with
$(\beta_1,\beta_2)=(0.9,0.95)$, weight decay 0.1, and gradient clipping
at 1.0. The learning-rate schedule advances with actual cumulative
tokens: 3\% warmup to $8\times10^{-5}$ followed by cosine decay to
$8\times10^{-6}$.
Each run executes 600 optimizer updates, saving and evaluating a
checkpoint every 100 updates. T90(obs.) uses the first of these six
checkpoints reaching 90\% joint accuracy, without interpolation.

\paragraph{Joint accuracy.}
All configurations use the same 1,000 validation problems, with 500
from each difficulty tier. Each problem receives one sample at
temperature 1 and top-$p=0.95$ under a fixed evaluation seed. The context
limit is 65,536 tokens; the output allowance subtracts the actual
prompt length. The native verifier requires both a valid process and
an executable, goal-reaching final plan, without requiring a shortest
or reference-matching path. Mixed+label receives one predetermined
strategy condition per problem: 250 BFS and 250 DFS labels within each
difficulty tier. A fixed-seed shuffle followed by alternating labels
sets this assignment before generation. Its 98.5\% accuracy is
985 successes among 1,000 conditional generations, with label adherence
tracked separately.

\paragraph{Dual-mode coverage.}
On 200 fixed problems, BFS-only, DFS-only, and Mixed each generate eight
unconditioned samples using the question and \texttt{<think>} prefix.
No strategy label or constrained decoding restricts their output mode.
Sampling uses temperature 1 and top-$p=0.95$. A problem is covered when
the generated samples include one jointly correct trace classified
exclusively as BFS and another classified exclusively as DFS.
Strategy is determined from the generated trajectory.
Mixed+label instead generates four samples
under each explicit strategy condition. The total sample budget is
thus eight in both protocols, with the strategy specified at inference
time only for Mixed+label. Coverage is computed from the observed
trajectory strategies.

The mixed configurations each cover 189 of 200 problems (94.5\%).
The single-strategy configurations cover zero problems under the same
unconditioned sampling budget; these observations do not assign zero
probability to the unobserved mode. Cov.@8 measures the observation of
both correct modes within eight samples.

\paragraph{Inference-time label removal.}
Evaluating the final Qwen3-4B Mixed+label checkpoint on the 1,000 validation
problems with the \texttt{STRATEGY} line removed (one sample per problem,
temperature $1$, top-$p=0.95$) yields 979/1,000 joint successes (97.9\%, vs.\
985/1,000 with labels; $-0.60$ pp, difficulty-stratified bootstrap 95\% CI
[$-1.50$, $+0.30$], exact McNemar $p=0.2863$; 971 shared successes, 7 shared
failures, 8 unlabeled-only successes, and 14 labeled-only successes), with
983/1,000 legal processes, 996/1,000 valid plans, 496/500 Easy successes,
483/500 Medium successes, and zero truncations. Among the 979 jointly correct
unlabeled traces, 960 (98.06\%) follow BFS exclusively, 19 (1.94\%) are
compatible with both strategies, and none follow DFS exclusively; this
single-sample test characterizes observed strategy preference rather than
dual-mode coverage, and we do not further verify the causes of this BFS bias.

\begin{table}[t]
    \centering
    \setlength{\tabcolsep}{6pt}\scalebox{0.9}{
    \begin{tabular}{@{}cccc@{}}
        \toprule
        Tokens (M) & BFS-only & DFS-only & Mixed \\
        \midrule
        100 & 15.4 & 66.6 & 33.0 \\
        200 & 71.6 & 86.6 & 79.8 \\
        300 & 83.4 & 93.4 & 92.2 \\
        400 & 88.4 & 93.8 & 90.8 \\
        500 & 98.4 & 95.2 & 96.4 \\
        600 & 98.4 & 97.2 & 98.6 \\
        \bottomrule
    \end{tabular}}
    \caption{Post-hoc joint success rate (\%, 500 problems) on the Medium validation subset across all six saved checkpoints for the three unlabeled supervision conditions. All runs initialize from Qwen3-4B-Base on 20K training problems ($\approx 600$M total token budget) and evaluate one sample per problem at temperature $1$ and top-$p=0.95$.}
    \label{tab:w1_medium}
\end{table}

\paragraph{Medium-subset trajectory across token budgets.}
Table~\ref{tab:w1_medium} reports a post-hoc difficulty-stratified breakdown across all six saved checkpoints (approximate cumulative budgets in millions of tokens, without additional training or generation) on the same 500 Medium validation problems for the three unlabeled conditions (BFS-only, DFS-only, and Mixed). Requiring both a valid search process and a correct final \texttt{PLAN}, Mixed tracks between the two single-mode baselines at 100M--200M tokens and reaches 92.2\% by 300M tokens, showing that Mixed can attain high joint success below the terminal budget.

\subsection{Scaling Experiments}
\label{app:scaling-experiments}

\paragraph{Data and budgets.}
The nested subsets preserve the difficulty and structural proportions of
the original pool. Each problem has two traces, one from each forward
search strategy. Inputs contain no strategy label. The mixture has equal
sequence counts; token proportions differ with trace lengths.
Table~\ref{tab:symbolic-training-budgets} gives the budgets for 10K,
20K, and 40K QA traces, corresponding to 5K, 10K, and 20K distinct
problems. Its token counts include input and target tokens, while
Updates and Warmup count optimizer steps over the complete six-epoch
schedule. At a fixed epoch, changing problem count also changes cumulative
training tokens and optimizer updates.
Computational cost varies across model sizes at a fixed token budget.

\begin{table}[t]
    \centering
    \setlength{\tabcolsep}{3pt}\scalebox{1}{
    \begin{tabular}{@{}lrrr@{}}
        \toprule
        Traces & Tokens/epoch & Updates & Warmup \\
        \midrule
        10K & 22.91M & 150 & 5 \\
        20K & 46.02M & 300 & 9 \\
        40K & 93.72M & 600 & 18 \\
        \bottomrule
    \end{tabular}}
    \caption{Training budgets shared across model sizes. Traces counts
    QA pairs (two per distinct problem). Tokens/epoch counts input and
    target tokens in one pass, excluding padding. Updates is the total
    number of optimizer steps over six epochs; Warmup counts the initial
    optimizer steps used to increase the learning rate. K and M denote
    thousands and millions.}
    \label{tab:symbolic-training-budgets}
\end{table}

\paragraph{Optimization.}
We use full-parameter BF16 SFT, global batch size 400.
Adam uses $(\beta_1,\beta_2)=(0.9,0.95)$, weight decay 0.1, and gradient
clipping at 1.0. The learning rate warms up to $8\times10^{-5}$ over 3\%
of updates, rounded up, and follows a cosine decay to $8\times10^{-6}$
over the full six-epoch run. All checkpoints belong to this same schedule.
The loss averages supervised assistant tokens, including the search trace
and final plan, with prompt and padding positions masked.

The context limit is 65,536 tokens, using YaRN with factor 2 from 32,768.
Sequences remain complete. Training uses 8 H20 GPUs.
Dynamic microbatch token limits are 32,768 for 0.6B/4B and 16,384 for 8B/14B.

\paragraph{Evaluation.}
The fixed validation split contains 500 problems from each of two difficulty
tiers. We sample one response per problem with temperature 1 and
 top-$p=0.95$. The generation limit is 65,536 minus the actual input length.
Incorrect and truncated responses remain in the denominator, and truncation
is reported separately. Any legal goal-reaching plan is accepted without
requiring a shortest or reference-matching path. Figures show 95\% Wilson
intervals for individual proportions. These intervals do not separately
measure training-seed variation or repeated-generation variation.

\subsection{An Actual QA Example}
\label{app:qa-example}

\paragraph{Model and generation settings.}
We show a single saved evaluation response from the epoch-6
Qwen3-4B-Base checkpoint, fine-tuned on 20K distinct problems with
40K forward BFS/DFS QA pairs, equally divided by strategy.
The response was generated with temperature $1$, top-$p=0.95$,
and a 65,536-token context limit.
No system message was used. The assistant was prefilled with
\texttt{<think>} followed by a newline; the response box includes
this prefix and the complete model continuation.

\begin{tcolorbox}[breakable,title={Prompt (user)},fonttitle=\bfseries\small,
    colback=black!2,colframe=black!50,colbacktitle=black!8,coltitle=black,
    boxrule=0.4pt,arc=1mm,left=6pt,right=6pt,top=4pt,bottom=4pt]
\begin{lstlisting}[basicstyle=\ttfamily\small,columns=fullflexible,keepspaces=true,breaklines=true]
RESOURCES A B C D E F G H
INITIAL B:1 D:1 E:1
GOAL H:1
RULES
r0: A:1 -> G:1
r1: F:1 G:1 -> A:1
r2: A:1 D:1 -> H:1
r3: A:1 -> H:1
r4: A:1 D:1 -> G:1
r5: D:1 -> C:1
r6: A:1 F:1 -> G:1
r7: B:1 -> G:1
r8: G:1 -> A:1
r9: B:1 -> E:1
\end{lstlisting}
\end{tcolorbox}

\begin{tcolorbox}[breakable,title={Model response (complete)},fonttitle=\bfseries\small,
    colback=black!2,colframe=black!50,colbacktitle=black!8,coltitle=black,
    boxrule=0.4pt,arc=1mm,left=6pt,right=6pt,top=4pt,bottom=4pt]
\begin{lstlisting}[basicstyle=\ttfamily\small,columns=fullflexible,keepspaces=true,breaklines=true]
<think>
EXPAND n0 STATE B:1 D:1 E:1
NEW n1 VIA r5 STATE B:1 C:1 E:1
NEW n2 VIA r7 STATE D:1 E:1 G:1
NEW n3 VIA r9 STATE D:1 E:2
CLOSE n0
EXPAND n1 STATE B:1 C:1 E:1
NEW n4 VIA r7 STATE C:1 E:1 G:1
NEW n5 VIA r9 STATE C:1 E:2
CLOSE n1
EXPAND n4 STATE C:1 E:1 G:1
NEW n6 VIA r8 STATE A:1 C:1 E:1
CLOSE n4
EXPAND n6 STATE A:1 C:1 E:1
DUP n4 VIA r0
NEW n7 VIA r3 STATE C:1 E:1 H:1
CLOSE n6
EXPAND n7 STATE C:1 E:1 H:1
GOAL n7
</think>

PLAN r5 r7 r8 r3
END
\end{lstlisting}
\end{tcolorbox}

\subsection{Mode Confusion}
\label{app:mode-confusion}

The declared-graph diagnostic checks the sequence of expansions implied
by the nodes and parent links written by the model. When the two
schedulers prescribe different choices, we record the expansion's
strategy if it matches one of them. A trajectory counts as switched
when both exclusive choices appear. The diagnostic can continue after
a semantic error while the declared graph is still structurally
interpretable. Table~\ref{tab:bfs_dfs_confusion} uses this diagnostic
for the 40K-trace epoch-based runs. Each checkpoint has 1,000 responses,
and a switched response counts once. The total across four sizes
declines from 43 at epoch one to one at epoch five. These checkpoints
and the 4B matched-token comparison in Appendix~\ref{app:matched-replay}
come from different training schedules.

\begin{figure}[t]
    \centering
    \includegraphics[width=\columnwidth]{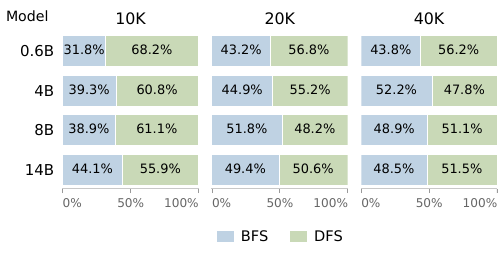}
    \caption{Marginal BFS and DFS proportions (\%) among identifiable
    successful validation trajectories across 10K, 20K, and 40K training
    trajectories. Each bar uses only the successful, single-mode subset
    of 1,000 responses. Subset counts are given in the text.}
    \label{fig:marginal-mode-ratio}
\end{figure}

For Figure~\ref{fig:marginal-mode-ratio}, the numbers of identifiable
single-mode successful trajectories (out of 1,000 validation problems)
for 0.6B, 4B, 8B, and 14B are 280, 530, 512, and 655 at 10K traces;
650, 796, 948, and 950 at 20K traces; and 851, 972, 963, and 972 at 40K
traces.

\begin{table}[t]
\centering
\setlength{\tabcolsep}{3pt}\scalebox{0.9}{
\begin{tabular}{@{}ccccc@{}}
\toprule
Model & Epoch & Acc. (\%) & Confusion & \shortstack{Succ. Conf.} \\
\midrule
\multirow{3}{*}{0.6B}
& 1 & 17.60 & 12 & 0 \\
& 3 & 64.50 & 3 & 0 \\
& 5 & 84.10 & 1 & 0 \\
\midrule
\multirow{3}{*}{4B}
& 1 & 35.90 & 3 & 0 \\
& 3 & 93.00 & 0 & 0 \\
& 5 & 98.00 & 0 & 0 \\
\midrule
\multirow{3}{*}{8B}
& 1 & 46.00 & 14 & 0 \\
& 3 & 89.90 & 1 & 1 \\
& 5 & 98.00 & 0 & 0 \\
\midrule
\multirow{3}{*}{14B}
& 1 & 9.50 & 14 & 1 \\
& 3 & 92.80 & 0 & 0 \\
& 5 & 99.00 & 0 & 0 \\
\bottomrule
\end{tabular}}
\caption{Results on 1,000 validation problems per checkpoint after
training on 40K trajectories. Acc. is the
percentage of responses with both a valid search and a correct final
answer. Succ. Conf. is the subset with a correct final answer.}
\label{tab:bfs_dfs_confusion}
\end{table}

\section{First-Error Replay Analysis}
\label{app:w3_methods}

\paragraph{Sample structure.}
The training set contains 20K distinct problems, each with one BFS and
one DFS trajectory, giving 40K QA pairs. We analyze the epoch-1 through
epoch-6 checkpoints of Qwen3-0.6B, 4B, 8B, and 14B. Each checkpoint
produces one response to each of the same 1,000 validation problems,
with 500 easy and 500 medium problems. Generation uses temperature 1,
top-$p=0.95$, a fixed evaluation seed, a 65,536-token context, and an
output budget equal to the context limit minus input length. The 24K
responses share 1,000 underlying problems; records across models and
epochs for the same problem, and events within a response, are related.
All rates below are pooled descriptive summaries of these records.

\paragraph{Valid-prefix opportunities.}
Replay accepts any legal frontier schedule and independently checks
the final plan. Successful responses contribute events through completion.
Failed responses contribute events through and including the first
detectable error. For expected event type $e$, let $O_e$ be the number
of these opportunities and $D_e$ the number at which the response first
fails. The empirical first-error rate is
\begin{equation}
    \widehat{\mathrm{Err}}(e)=\frac{D_e}{O_e}.
    \label{eq:w3_event_error}
\end{equation}
This is the first-error proportion conditional on an otherwise valid
prefix and an expected event of type $e$. For example, the
\texttt{NEW} rate of 0.2972\% corresponds to 29.72 first errors per
10,000 such opportunities. Observation of each response ends at its
first error or successful completion.

\paragraph{Expected-event classification.}
The replay phase and expected successor sequence determine event type
before the current output is read. When \texttt{NEW} is expected but
\texttt{DUP} is emitted, both the opportunity and its first error are
assigned to \texttt{NEW}; the emitted type is recorded separately.
Expected and emitted types can also differ because of rule ordering,
omitted successors, or malformed output. We identify a same-rule
\texttt{NEW}/\texttt{DUP} confusion when the verifier has matched the
same rule and the emitted type disagrees with its expected type.
There are 1,661 \texttt{NEW}$\rightarrow$\texttt{DUP} and 1,941
\texttt{DUP}$\rightarrow$\texttt{NEW} confusions, totaling
$3,602/5,318=67.73\%$ of first errors. The broader count of expected
versus emitted type mismatches is 3,881. Together, expected
\texttt{NEW} and \texttt{DUP} contribute 1,742,456 opportunities
(57.05\% of all event types) and 4,904 first errors (92.22\%).

\paragraph{Final-plan checks.}
At the \texttt{PLAN} line, the verifier independently executes the
complete rule sequence, checking rule existence, resource availability,
and goal attainment. It accepts executable goal-reaching sequences
regardless of agreement with the reference path or path length.
Thus, a \texttt{PLAN} opportunity is a sequence-level check, whereas
\texttt{NEW} and \texttt{DUP} each record a single successor.
The 218 \texttt{PLAN} first errors comprise 213 cases (97.71\%) of
insufficient resources, two unmet goals, one syntax error, one unknown
rule, and one sequence exceeding the 512-step verification limit.
Insufficient resources means that the task inventory at some step of
final-plan execution cannot supply the inputs of the next rule. The
preceding search events have passed replay, so these cases locate a
failure in final-plan executability. Our analysis of critical reasoning
decisions focuses on \texttt{NEW} and \texttt{DUP}, which govern state
registration and recognition within the search process.

\subsection{Matched-Token Replay}
\label{app:matched-replay}
The matched-token comparison uses Qwen3-4B, training seed 17, and one
unconditioned response per validation problem at temperature 1 and
top-$p=0.95$. We replay the same 1,000 problems (500 Easy, 500 Medium)
at six checkpoints in each of the three training conditions. All
18,000 generations are retained, including truncations. Every joint
failure in this replay has a localized first error.

\paragraph{Denominators and uncertainty.}
At an expansion where the BFS and DFS schedulers prescribe different
nodes, a choice matching exactly one scheduler is informative.
A response contributes to the switch denominator when it makes at
least two informative choices before the first semantic error. We
record a switch when both strategies appear in that subsequence.
The declared-graph diagnostic can also count choices after the first
semantic error, stopping when the graph becomes structurally
uninterpretable. It finds one Mixed switch at each of 400M and 500M
tokens (Table~\ref{tab:matched-switch}), while the valid-prefix
diagnostic finds none. Switch intervals are pointwise Wilson 95\%
intervals.

The broad and strict first-error shares both use all localized first
errors as their denominator. The opportunity rate uses every expected
\texttt{NEW}/\texttt{DUP} event up to completion or the first error,
including the error opportunity. Rate intervals resample validation
problems within difficulty strata 10,000 times. The same problems
appear at each checkpoint, and these intervals estimate
within-checkpoint problem-sampling uncertainty for the observed
training run. Table~\ref{tab:matched-replay-counts} gives both
denominators at every checkpoint. At 600M, the strict shares are 7/12 for BFS-only, 16/23 for
DFS-only, and 6/8 for Mixed, with Wilson 95\% intervals
[31.95, 80.67]\%, [49.13, 84.40]\%, and [40.93, 92.85]\%.

\paragraph{Exploratory accuracy matching.}
Using existing validation accuracies, we selected all Mixed--single-strategy
checkpoint pairs where both joint accuracies lie between 70\% and 96\% and
differ by at most two pp, before inspecting the new replay statistics.
Six pairs meet this criterion in Table~\ref{tab:matched-replay-counts}.
The Mixed 200M checkpoint matches BFS-only at 300M. Mixed at 300M
and 400M each match DFS-only at both 300M and 400M, and Mixed at
400M also matches BFS-only at 400M. Opportunity rates also depend on trace lengths and event composition, so the
comparisons describe the observed runs.

\paragraph{Example with distinct outcomes.}
In one Mixed 100M response, replay stops at event line 28 because rule
\texttt{r7} requires a \texttt{NEW} event and the model emits \texttt{DUP}.
The independent plan check still succeeds. This response contributes one
strict confusion and one plan-only success.

\begin{table*}[t]
\centering
\small
\setlength{\tabcolsep}{5pt}
\begin{tabular}{@{}lrrrrrr@{}}
\toprule
Training & Tokens (M) & Joint (/1,000) & Broad/errors & Strict/errors & Opportunities & Strict/10K [95\% CI] \\
\midrule
BFS-only & 100.001 & 237 & 718/763 & 637/763 & 41,178 & 154.694 [135.14, 178.04] \\
BFS-only & 200.007 & 809 & 152/191 & 100/191 & 108,871 & 9.185 [7.43, 11.07] \\
BFS-only & 300.023 & 879 & 106/121 & 59/121 & 122,887 & 4.801 [3.61, 6.13] \\
BFS-only & 400.005 & 927 & 68/73 & 43/73 & 130,992 & 3.283 [2.34, 4.28] \\
BFS-only & 500.017 & 985 & 11/15 & 7/15 & 143,660 & 0.487 [0.15, 0.87] \\
BFS-only & 600.002 & 988 & 10/12 & 7/12 & 144,751 & 0.484 [0.15, 0.86] \\
DFS-only & 100.003 & 712 & 277/288 & 255/288 & 28,863 & 88.348 [77.69, 99.60] \\
DFS-only & 200.001 & 898 & 85/102 & 64/102 & 38,138 & 16.781 [12.90, 21.10] \\
DFS-only & 300.002 & 948 & 47/52 & 33/52 & 41,608 & 7.931 [5.37, 10.72] \\
DFS-only & 400.004 & 950 & 48/50 & 29/50 & 41,868 & 6.927 [4.59, 9.58] \\
DFS-only & 500.003 & 963 & 32/37 & 29/37 & 42,705 & 6.791 [4.47, 9.26] \\
DFS-only & 600.000 & 977 & 20/23 & 16/23 & 45,699 & 3.501 [1.94, 5.22] \\
Mixed & 100.004 & 374 & 615/626 & 564/626 & 30,569 & 184.501 [162.03, 208.95] \\
Mixed & 200.008 & 862 & 105/138 & 83/138 & 75,892 & 10.937 [8.70, 13.43] \\
Mixed & 300.000 & 948 & 45/52 & 25/52 & 89,665 & 2.788 [1.74, 3.91] \\
Mixed & 400.013 & 934 & 54/66 & 37/66 & 94,474 & 3.916 [2.71, 5.27] \\
Mixed & 500.009 & 981 & 15/19 & 11/19 & 92,491 & 1.189 [0.55, 1.95] \\
Mixed & 600.001 & 992 & 7/8 & 6/8 & 90,277 & 0.665 [0.21, 1.25] \\
\bottomrule
\end{tabular}
\caption{First-error composition and opportunity-normalized strict
confusion across all 18 matched-budget checkpoints. Broad counts any
first error at an expected \texttt{NEW}/\texttt{DUP} event. Strict counts
reversed event types with the expected rule.}
\label{tab:matched-replay-counts}
\end{table*}

\begin{table*}[t]
\centering
\small
\setlength{\tabcolsep}{4pt}
\begin{tabular}{@{}lrrrrrrr@{}}
\toprule
& & \multicolumn{3}{c}{Valid prefix before first error}
    & \multicolumn{2}{c}{Model-declared graph} & \\
\cmidrule(lr){3-5}\cmidrule(lr){6-7}
Training & Tokens (M) & Reached & Switched/eligible & 95\% CI (\%)
    & Switched/eligible & Stops & Truncated \\
\midrule
BFS-only & 100 & 877 & 0/751 & [0.00, 0.51] & 0/947 & 101 & 32 \\
 & 200 & 973 & 0/954 & [0.00, 0.40] & 0/968 & 10 & 0 \\
 & 300 & 976 & 0/960 & [0.00, 0.40] & 0/968 & 3 & 1 \\
 & 400 & 981 & 0/967 & [0.00, 0.40] & 0/968 & 1 & 0 \\
 & 500 & 981 & 0/967 & [0.00, 0.40] & 0/968 & 0 & 0 \\
 & 600 & 981 & 0/968 & [0.00, 0.40] & 0/968 & 0 & 0 \\
\addlinespace[4pt]
DFS-only & 100 & 949 & 0/894 & [0.00, 0.43] & 0/947 & 14 & 16 \\
 & 200 & 970 & 0/932 & [0.00, 0.41] & 0/947 & 6 & 4 \\
 & 300 & 977 & 0/942 & [0.00, 0.41] & \textbf{1}/947 & 1 & 1 \\
 & 400 & 971 & \textbf{1}/935 & [0.02, 0.60] & \textbf{1}/947 & 0 & 0 \\
 & 500 & 979 & 0/944 & [0.00, 0.41] & \textbf{1}/947 & 2 & 1 \\
 & 600 & 979 & 0/943 & [0.00, 0.41] & 0/947 & 0 & 0 \\
\addlinespace[4pt]
Mixed & 100 & 909 & 0/819 & [0.00, 0.47] & 0/957 & 56 & 32 \\
 & 200 & 979 & 0/946 & [0.00, 0.40] & 0/956 & 4 & 0 \\
 & 300 & 980 & 0/952 & [0.00, 0.40] & 0/953 & 2 & 0 \\
 & 400 & 981 & 0/959 & [0.00, 0.40] & \textbf{1}/962 & 2 & 1 \\
 & 500 & 981 & 0/957 & [0.00, 0.40] & \textbf{1}/957 & 3 & 0 \\
 & 600 & 980 & 0/955 & [0.00, 0.40] & 0/956 & 0 & 0 \\
\bottomrule
\end{tabular}
\caption{Strategy switches on 1,000 problems per checkpoint. Reached and
eligible require at least one and two strategy-identifying expansions,
respectively. Bold counts mark switches. Wilson intervals cover the
valid-prefix rate. Stops count unreadable declared graphs, and Truncated
counts responses at the output length limit.}
\label{tab:matched-switch}
\end{table*}

\section{Local Assistance Evaluation}
\label{app:local-assistance}

\paragraph{Case selection and continuation.}
The intervention pool is drawn from the epoch-2 validation responses
in the 40K-trace training setting. We retain every case whose first
replay error is a same-rule \texttt{NEW}/\texttt{DUP} type confusion,
using the rule-matching criterion in Appendix~\ref{app:w3_methods}.
The resulting pool contains 495, 98, 88, and 86 cases for 0.6B, 4B,
8B, and 14B, respectively. The 767 model--problem cases share 552
distinct source problems. Each case uses the valid prefix immediately
before the first error, and the same original checkpoint across A, B,
and C. Four continuations are sampled per condition, giving 9,204
continuations in total. All continuations are evaluated by the
environment for process legality and final-plan correctness.

\paragraph{Estimands and aggregation.}
For case $i$, condition $k\in\{A,B,C\}$, and continuation
$s\in\{1,\ldots,4\}$, let $R_{iks}$ be the applicable outcome
indicator. Joint success requires legal replay of the retained prefix
and continuation plus a valid plan. Plan-only success extracts the
final \texttt{PLAN} and executes it from the original initial inventory
against the same rule, resource, and goal checks.
The case-level success estimate and condition contrast are
\begin{equation}
    \widehat v_{ik}=\frac14\sum_{s=1}^{4}R_{iks},
    \qquad
    \widehat\delta_i^{k\ell}=\widehat v_{ik}-\widehat v_{i\ell}.
    \label{eq:case-level-estimands}
\end{equation}
For a set of cases $\mathcal I$, we report
\begin{equation}
    \widehat v_k=\frac{1}{|\mathcal I|}
    \sum_{i\in\mathcal I}\widehat v_{ik},
    \qquad
    \widehat\Delta_{k-\ell}=\frac{1}{|\mathcal I|}
    \sum_{i\in\mathcal I}\widehat\delta_i^{k\ell}.
    \label{eq:sample-mean-estimands}
\end{equation}
Each model-specific estimate averages over that model's selected failure
pool. The pooled estimate gives equal weight to all 767 cases, so model
weights follow their case counts. These estimands describe recovery
from the selected valid prefixes under the three assistance conditions.
To connect these estimates to
Equation~\ref{eq:local-assistance-value}, let $h_i$ and $\theta_i$
denote case $i$'s valid prefix and original checkpoint, and let $I_k$
be the assistance procedure in condition $k$. The target of each
assisted-versus-unassisted contrast is
\begin{equation}
    \Delta_{k-A}
    =\frac{1}{|\mathcal I|}\sum_{i\in\mathcal I}
    \Delta_{I_k}^{\theta_i}(h_i),
    \qquad k\in\{B,C\}.
    \label{eq:target-contrast}
\end{equation}
Thus, $\widehat\Delta_{C-A}$ estimates the mean local assistance
value over the selected cases. The C$-$B contrast estimates the
difference between the mean assistance values of the two procedures.

\paragraph{Uncertainty and paired comparisons.}
The 95\% confidence intervals use 10,000 bootstrap resamples clustered
by source problem. Each sampled problem carries all its model-specific
cases, conditions, and continuation samples together. Paired condition
comparisons use two-sided source-problem-level sign-flip tests with
100,000 Monte Carlo draws. Holm correction \citep{holm1979simple}
covers the three pairwise joint-success comparisons for each model
and the pooled summary (15 comparisons).
Table~\ref{tab:local-assistance} reports the Holm-adjusted joint
C$-$A tests. Table~\ref{tab:assistance-intervals} gives pointwise
intervals for the success rates under each condition.

\paragraph{Additional contrasts.}
Pooled joint B$-$A and C$-$B gains are 28.00 and 0.85 pp, respectively.
The plan-only C$-$B contrast is $-0.07$ pp [$-1.32$, $+1.22$].
The counts of continuations with a valid plan and an invalid process
are 2,166 (A), 1,326 (B), and 1,298 (C). Of 3,068 continuations per
condition, 2,673, 2,692, and 2,690 final plans pass independent execution
under A, B, and C, respectively. In A, 507 continuations pass both checks. The contrasts condition on
the chosen first-error prefixes.

\begin{table}[t]
\centering
\small
\setlength{\tabcolsep}{2.5pt}
\begin{tabular}{@{}lrrr@{}}
\toprule
Model & Unassisted & Type hint & Full repair \\
\midrule
\multicolumn{4}{@{}l}{Joint trace-and-plan success} \\
0.6B & [8.18, 12.27] & [28.59, 35.96] & [28.74, 36.26] \\
4B & [21.43, 34.95] & [54.85, 71.68] & [58.16, 74.49] \\
8B & [17.90, 31.83] & [56.53, 73.86] & [61.93, 78.12] \\
14B & [23.84, 39.83] & [64.83, 80.23] & [61.92, 78.49] \\
Pooled & [14.54, 18.57] & [41.47, 47.54] & [42.36, 48.47] \\
\midrule
\multicolumn{4}{@{}l}{Independent final-plan accuracy} \\
Pooled & [84.83, 89.30] & [85.55, 89.80] & [85.45, 89.80] \\
\bottomrule
\end{tabular}
\caption{Pointwise 95\% confidence intervals (\%) for the success
rates in Table~\ref{tab:local-assistance}, clustered by source problem.}
\label{tab:assistance-intervals}
\end{table}

\section{Natural Language Benchmark Setup and Statistical Protocol}
\label{app:crn-nl-details}

\subsection{Models and Data}
\paragraph{Benchmarks.}
GPQA-Diamond comprises 198 graduate-level multiple-choice questions. For HLE, we filter the test split for text-only questions without accompanying images (2,158 eligible problems) and uniformly sample 100 problems without replacement under a fixed random seed. Both models are evaluated on the exact same problem set and reference answers, consisting of 80 short-answer questions and 20 multiple-choice questions with original option ordering preserved.

\paragraph{Inference configurations.}
DeepSeek V4.1 Flash \citep{deepseekai2026deepseekv41flash} and Qwen3.8-27B \citep{qwen2026qwen38} are evaluated in thinking mode with top-$p=0.95$, allowing up to 65,536 generated tokens per response.

\subsection{Answer Verification and Error Localization}
\paragraph{Answer grading.}
Explicit final answers are first verified via rule-based matching against the reference answer and option mapping; when rule extraction fails or returns an initial mismatch on a non-empty visible response, the answer is adjudicated by \texttt{gpt-6.2-sol} (with \texttt{gpt-6.1-sol} as fallback under the identical prompt in the event of technical or protocol failures). Truncated generations are tracked separately and counted as end-to-end failures.

\paragraph{Error localization and independent verification.}
Across all model--benchmark pairs, we use GPT-6-Astra as a unified automated localizer: for each incorrect trajectory, it inspects the model's reasoning trace alongside the problem statement, locates the exact text spans of independent or downstream reasoning errors, and generates a minimal replacement step for each repairable position. Localization inputs include only the problem statement and the model's original trajectory; reference answers are restricted exclusively to answer grading. To verify the reliability of GPT-6-Astra's localization and repair generation, we conduct an independent stratified audit using \texttt{gpt-6.1-sol} on 48 randomly sampled nodes (12 per model--benchmark stratum; Table~\ref{tab:astra-verification}). Across the 48 audited nodes, \texttt{gpt-6.1-sol} confirms 42 localized error positions as genuine errors (87.50\%), and among these 42 confirmed errors, validates 33 proposed local repairs as accurate (78.57\%). Weighting the four sampling strata by their respective sizes in the full 3,197-node pool yields a stratum-weighted localization confirmation rate of 90.15\% and a repair validity rate of 84.54\%.

\begin{table}[t]
\centering
\setlength{\tabcolsep}{3.5pt}\scalebox{0.75}{
\begin{tabular}{@{}llrr@{}}
\toprule
Model & Benchmark & \shortstack{Error loc.\\confirmed} & \shortstack{Local repair\\confirmed} \\
\midrule
Qwen3.8-27B & GPQA-D & 11/12 (91.67\%) & 6/11 (54.55\%) \\
Qwen3.8-27B & HLE & 11/12 (91.67\%) & 10/11 (90.91\%) \\
DeepSeek & GPQA-D & 9/12 (75.00\%) & 8/9 (88.89\%) \\
DeepSeek & HLE & 11/12 (91.67\%) & 9/11 (81.82\%) \\
\midrule
Sample total & --- & 42/48 (87.50\%) & 33/42 (78.57\%) \\
Stratum-weighted & --- & 90.15\% & 84.54\% \\
\bottomrule
\end{tabular}}
\caption{Independent verification of GPT-6-Astra error localization and local repair quality by \texttt{gpt-6.1-sol} across 48 stratified random samples (12 per model--benchmark stratum). Repair confirmation uses confirmed error nodes as the denominator; stratum-weighted estimates weight the four strata by their sizes among all 3,197 localized errors.}
\label{tab:astra-verification}
\end{table}

\subsection{Atomic Probing and Local Continuation Protocol}
\paragraph{Atomic probe generation and evaluation.}
To isolate reproducible step-level bottlenecks among all 3,197 localized error positions (1,937 from Qwen3.8-27B and 1,260 from DeepSeek V4.1 Flash across the 4 baseline rollouts on 198 GPQA-Diamond and 100 text-only HLE problems), \texttt{gpt-6.2-sol} generates one standalone open-ended short-answer probe per error position based on the localized error span, rationale, minimal repair, upstream dependencies, and surrounding context identified by GPT-6-Astra. Each probe supplies the necessary local premises to test only the minimal knowledge or reasoning operation at that error position, together with a canonical reference answer and equivalent aliases. Both Qwen3.8-27B and DeepSeek V4.1 Flash independently answer all 3,197 probes 4 times each (25,576 probe responses in total) in thinking mode with a 65,536-token generation limit, formatting their final line as \texttt{FINAL: <answer>} (Qwen3.8-27B runs locally with temperature $1.0$, top-$p=0.95$, and top-$k=20$; DeepSeek V4.1 Flash uses \texttt{high} reasoning effort and top-$p=0.95$).

\paragraph{Probe grading and recurrence.}
Rule-based extraction compares the response with the canonical
answer and aliases. DeepSeek adjudicates semantic equivalence for
visible responses that fail extraction or rule matching, against the
fixed reference answer. The recurrence count $k$ uses the four attempts
by the original model. Truncations, responses without a visible final
answer, and explicitly undetermined responses count as failed attempts
in the frozen grading. Technical failures are retried or re-adjudicated.

\paragraph{Local continuation.}
Condition A resumes generation from the original assistant prefix.
Condition C appends the local repair before resuming from that prefix.
Each site has one continuation per condition. Both conditions discard
all original text from the targeted span onward. The full localization
pool covers 3,197 errors on 95 problems. Of these, 16 lack a usable
repair, 698 repairs directly supply the answer, and five have an
incomplete A/C pair. The 2,478 eligible pairs span 93 problems:
406 pairs from 20 GPQA-Diamond problems and 2,072 pairs from 73 HLE
problems. Whole-problem baseline accuracy
(Table~\ref{tab:nl-baseline-acc}) describes the source tasks. The A
condition estimates success from the selected prefixes.

\begin{table}[t]
\centering
\setlength{\tabcolsep}{2.5pt}\scalebox{0.95}{
\begin{tabular}{@{}lcccc@{}}
\toprule
Model & \shortstack{GPQA-\\Diamond} & \shortstack{HLE\\(text)} & \shortstack{Macro\\avg.} & \shortstack{Pooled\\by trace} \\
\midrule
Qwen3.8-27B & \shortstack{89.65\%\\{\scriptsize (710/792)}} & \shortstack{24.75\%\\{\scriptsize (99/400)}} & 57.20\% & \shortstack{67.87\%\\{\scriptsize (809/1,192)}} \\
\addlinespace[2pt]
DeepSeek & \shortstack{91.04\%\\{\scriptsize (721/792)}} & \shortstack{34.75\%\\{\scriptsize (139/400)}} & 62.89\% & \shortstack{72.15\%\\{\scriptsize (860/1,192)}} \\
\bottomrule
\end{tabular}}
\caption{Baseline accuracy across four rollouts per problem on GPQA-Diamond and the 100-problem text-only HLE subset for Qwen3.8-27B and DeepSeek V4.1 Flash (DeepSeek).}
\label{tab:nl-baseline-acc}
\end{table}

\subsection{Node-Level Repair and Probe Recurrence}
\label{app:probe-repair}
We estimate repair gain as the mean of the paired C$-$A correctness
indicators, with each eligible node receiving equal weight. A node-level
linear regression estimates the slope in $k$, both without covariates
and with model-by-benchmark fixed effects. We compute pointwise 95\%
percentile intervals from 10,000 bootstrap samples of source problems,
clustered by benchmark and original problem ID. Each resample retains
the model, trajectory, and node records of the sampled problem together.

Table~\ref{tab:probe-bands} reports the five recurrence bands. The
full-sample gain is 3.79 pp, corresponding to 94 additional correct
continuations over 2,478 pairs (146 correct under A and 240 under C).
The $k\geq2$ group gains 4.88 pp [$+0.86$, $+10.56$], and
the $k<2$ group gains 3.59 pp [$+1.82$, $+5.66$].
Table~\ref{tab:crn-nl-tasks} retains the high-recurrence results for each
model and benchmark. The Qwen GPQA-Diamond repair audit confirms
6 of 11 repairs at confirmed errors (Table~\ref{tab:astra-verification}),
which warrants caution when interpreting this subgroup's large gain.
The upstream sensitivity analysis uses the localizer's flag and
includes 303 valid pairs with no flagged earlier error.
Table~\ref{tab:probe-slopes} reports the corresponding slopes.

\paragraph{Fixed node budgets.}
We compare selecting the 389 high-recurrence nodes with the expected
gain from selecting 389 eligible nodes uniformly without replacement.
We also match the observed numbers selected per model and benchmark,
or per original failed trajectory. The resamples are shared between
probe-based and random selection. For the model-by-benchmark quota
comparison, 9,996 draws contain all quota strata. The three pointwise
intervals for the gain difference span zero
(Table~\ref{tab:probe-node-random}).

\begin{table*}[t]
\centering
\small
\setlength{\tabcolsep}{4pt}
\begin{tabular}{@{}lrrrrrrr@{}}
\toprule
Subset & $k$ & Sites / problems & Pairs / problems & Correct A & Correct C &
C$-$A (pp) & 95\% CI (pp) \\
\midrule
All & 0 & 2,520 / 95 & 1,959 / 93 & 113 & 183 & $+3.57$ & [$+1.83$, $+5.68$] \\
 & 1 & 180 / 67 & 130 / 62 & 9 & 14 & $+3.85$ & [$-2.46$, $+10.74$] \\
 & 2 & 102 / 51 & 83 / 45 & 4 & 6 & $+2.41$ & [$-2.08$, $+7.69$] \\
 & 3 & 103 / 40 & 84 / 36 & 5 & 11 & $+7.14$ & [$+1.00$, $+15.15$] \\
 & 4 & 292 / 49 & 222 / 46 & 15 & 26 & $+4.95$ & [$-0.41$, $+12.35$] \\
GPQA-Diamond & 0 & 481 / 20 & 325 / 20 & 45 & 84 & $+12.00$ & [$+6.37$, $+18.27$] \\
 & 1 & 25 / 12 & 18 / 10 & 2 & 8 & $+33.33$ & [$+5.88$, $+64.29$] \\
 & 2 & 17 / 12 & 12 / 9 & 2 & 3 & $+8.33$ & [$+0.00$, $+23.08$] \\
 & 3 & 19 / 9 & 14 / 8 & 3 & 7 & $+28.57$ & [$+4.54$, $+66.67$] \\
 & 4 & 43 / 13 & 37 / 12 & 3 & 13 & $+27.03$ & [$+6.90$, $+50.00$] \\
HLE (text) & 0 & 2,039 / 75 & 1,634 / 73 & 68 & 99 & $+1.90$ & [$+0.45$, $+3.69$] \\
 & 1 & 155 / 55 & 112 / 52 & 7 & 6 & $-0.89$ & [$-6.12$, $+4.46$] \\
 & 2 & 85 / 39 & 71 / 36 & 2 & 3 & $+1.41$ & [$-3.08$, $+6.90$] \\
 & 3 & 84 / 31 & 70 / 28 & 2 & 4 & $+2.86$ & [$-2.82$, $+7.81$] \\
 & 4 & 249 / 36 & 185 / 34 & 12 & 13 & $+0.54$ & [$-3.40$, $+4.14$] \\
No flagged upstream error & 0 & 252 / 91 & 233 / 88 & 27 & 44 & $+7.30$ & [$+2.20$, $+13.02$] \\
 & 1 & 15 / 14 & 15 / 14 & 1 & 4 & $+20.00$ & [$+0.00$, $+42.86$] \\
 & 2 & 11 / 11 & 7 / 7 & 0 & 0 & $+0.00$ & [$+0.00$, $+0.00$] \\
 & 3 & 13 / 12 & 13 / 12 & 2 & 5 & $+23.08$ & [$+0.00$, $+50.00$] \\
 & 4 & 44 / 21 & 35 / 18 & 3 & 6 & $+8.57$ & [$-7.90$, $+24.39$] \\
\bottomrule
\end{tabular}
\caption{Node-level repair by probe failure count. The two problem
counts refer to all located sites and to valid pairs, respectively.
Problems can contribute to multiple bands. The upstream subset uses
the localizer's recorded flag.}
\label{tab:probe-bands}
\end{table*}

\begin{table*}[t]
\centering

\setlength{\tabcolsep}{6pt}
\begin{tabular}{@{}lrrr@{}}
\toprule
Subset & Pairs & Unadjusted slope [95\% CI] & Adjusted slope [95\% CI] \\
\midrule
All & 2,478 & $+0.42$ [$-0.90$, $+2.21$] & $+0.50$ [$-0.75$, $+2.00$] \\
GPQA-Diamond & 406 & $+3.95$ [$-0.97$, $+9.91$] & $+4.13$ [$-0.81$, $+10.18$] \\
HLE (text) & 2,072 & $-0.27$ [$-1.33$, $+0.65$] & $-0.22$ [$-1.26$, $+0.66$] \\
No flagged upstream error & 303 & $+1.07$ [$-2.87$, $+5.04$] & $+1.80$ [$-1.86$, $+5.10$] \\
Qwen3.8-27B & 1,582 & $+0.39$ [$-0.87$, $+1.98$] & $+0.40$ [$-0.83$, $+1.74$] \\
DeepSeek V4.1 Flash & 896 & $+1.23$ [$-1.31$, $+4.50$] & $+0.89$ [$-1.46$, $+4.00$] \\
Qwen / GPQA-D & 208 & $+5.96$ [$+0.07$, $+11.39$] & $+5.96$ [$+0.07$, $+11.39$] \\
Qwen / HLE & 1,374 & $-0.37$ [$-1.53$, $+0.56$] & $-0.37$ [$-1.53$, $+0.56$] \\
DeepSeek / GPQA-D & 198 & $+1.67$ [$-4.52$, $+11.98$] & $+1.67$ [$-4.52$, $+11.98$] \\
DeepSeek / HLE & 698 & $+0.49$ [$-1.24$, $+2.98$] & $+0.49$ [$-1.24$, $+2.98$] \\
\bottomrule
\end{tabular}
\caption{Slope of paired repair gain, in pp per
additional failed probe. The adjusted specification uses
model-by-benchmark fixed effects.}
\label{tab:probe-slopes}
\end{table*}

\begin{table*}[t]
\centering
\begin{tabular}{@{}lrrrr@{}}
\toprule
Random selection pool & Probe gain & Random gain & Difference & 95\% CI \\
\midrule
All eligible nodes & $+4.88$ & $+3.79$ & $+1.09$ & [$-2.31$, $+5.79$] \\
Model-by-benchmark quotas & $+4.88$ & $+3.60$ & $+1.29$ & [$-1.84$, $+4.71$] \\
Original-trajectory quotas & $+4.88$ & $+4.00$ & $+0.88$ & [$-1.31$, $+3.04$] \\
\bottomrule
\end{tabular}
\caption{Comparison with uniform node selection under the same
389-node repair budget. Gains and contrasts are in pp.}
\label{tab:probe-node-random}
\end{table*}

\subsection{Retrospective One-Repair Selection}
\label{app:one-repair-replay}
The archived replay starts from 555 baseline trajectories graded
incorrect, of which 511 were submitted for localization. The
whole-problem accuracy denominators in Table~\ref{tab:nl-baseline-acc}
cover all baseline rollouts, including truncations.
Before linking historical outcomes, we freeze the
candidate list using the existing localization and repairs, removing
sites without a usable local repair and sites marked as revealing the
final answer. The resulting pool contains 2,483 candidates across 330
trajectories and 93 original problems. Of these trajectories, 302 have
multiple candidates. Earliest refers to the first eligible location,
ordering reasoning before answer text and using the character offset
within each region. Probe selection takes the highest $k$, breaking
ties by that order. The decisions differ on 135 trajectories spanning
79 source problems. Uniform random selection is evaluated by the exact
within-trajectory mean over eligible candidates' recorded outcomes.

\paragraph{Outcomes and missingness.}
Both deterministic policies have a C outcome on all 330 trajectories
(60 earliest successes and 50 probe successes). One trajectory has a
missing C outcome for another eligible candidate, leaving 329
trajectories for the comparison with random selection. Complete
A/C outcomes for all three rules are available on 326 trajectories.
Table~\ref{tab:one-repair-flow} traces these denominators by benchmark
and model.
Each trajectory receives equal weight. Intervals use 10,000 paired
source-problem cluster bootstrap draws. On the 329 common trajectories,
the probe policy alone succeeds on 10 and the earliest policy alone on
20. Imputing either possible outcome for the missing C puts the
probe-minus-random contrast on all 330 trajectories between $+3.85$
and $+3.90$ pp. The probe-minus-earliest contrast on all 330 is
$-3.03$ pp [$-6.45$, $+0.28$]. In Table~\ref{tab:one-repair-by-task}, earliest selection has the highest
observed repaired accuracy on both benchmarks. On the complete A/C sample,
it also has the highest unassisted accuracy on both
(Table~\ref{tab:one-repair-ac}). Table~\ref{tab:one-repair-gains}
compares the paired gains directly. For the full sample of 326, probe
minus earliest is $+0.61$ pp, with a 95\% interval spanning zero.

\paragraph{Cost and interpretation.}
Skipping probes for the 28 single-candidate trajectories leaves 9,820
additional source-model calls consuming 25,958,469 generated tokens,
including thinking and visible answers. All strategies share
localization and repair costs. The probe rule's advantage over random
selection also includes the earliest-position tie breaker. This
historical analysis reuses the node-level problem pool and uses one A/C sample
per site. A prospective comparison can freeze the candidates and rules
on new problems, share repeated continuations for common sites, and
designate the earliest rule as the primary comparator under the same
one-repair budget.

\begin{table*}[t]
\centering
\small
\setlength{\tabcolsep}{6pt}
\begin{tabular}{@{}lrrrrrrr@{}}
\toprule
Subset & Problems & Eligible & Multiple & Disagree & Disagree problems &
Complete C & Complete A/C \\
\midrule
All eligible & 93 & 330 & 302 & 135 & 79 & 329 & 326 \\
Multiple sites & 92 & 302 & 302 & 135 & 79 & 301 & 298 \\
GPQA-Diamond & 20 & 60 & 56 & 27 & 16 & 59 & 56 \\
HLE (text) & 73 & 270 & 246 & 108 & 63 & 270 & 270 \\
Qwen / GPQA-D & 14 & 28 & 26 & 14 & 12 & 28 & 28 \\
Qwen / HLE & 56 & 145 & 132 & 64 & 43 & 145 & 145 \\
DeepSeek / GPQA-D & 15 & 32 & 30 & 13 & 9 & 31 & 28 \\
DeepSeek / HLE & 58 & 125 & 114 & 44 & 33 & 125 & 125 \\
\bottomrule
\end{tabular}
\caption{Trajectory counts for the one-repair replay. Disagreement
refers to probe and earliest choosing different sites. Complete outcomes
refer to the sample shared by all three policies.}
\label{tab:one-repair-flow}
\end{table*}

\begin{table*}[t]
\centering
\small
\setlength{\tabcolsep}{5pt}
\begin{tabular}{@{}lrrrrrr@{}}
\toprule
Subset & $n$ & Earliest & Random & Probe &
Probe$-$earliest [95\% CI] & Probe$-$random [95\% CI] \\
\midrule
GPQA-Diamond & 59 & 50.85 & 33.38 & 40.68 & $-10.17$ [$-25.72$, $+1.82$] & $+7.30$ [$-1.09$, $+14.73$] \\
HLE (text) & 270 & 11.11 & 6.46 & 9.63 & $-1.48$ [$-4.41$, $+1.42$] & $+3.17$ [$+0.61$, $+6.11$] \\
Qwen / GPQA-D & 28 & 50.00 & 32.54 & 39.29 & $-10.71$ [$-34.78$, $+8.57$] & $+6.74$ [$-6.44$, $+19.61$] \\
Qwen / HLE & 145 & 11.72 & 7.46 & 11.03 & $-0.69$ [$-4.86$, $+3.36$] & $+3.57$ [$+0.22$, $+7.35$] \\
DeepSeek / GPQA-D & 31 & 51.61 & 34.13 & 41.94 & $-9.68$ [$-25.00$, $+3.23$] & $+7.80$ [$-2.35$, $+18.31$] \\
DeepSeek / HLE & 125 & 10.40 & 5.29 & 8.00 & $-2.40$ [$-6.67$, $+1.61$] & $+2.71$ [$-1.75$, $+8.02$] \\
\bottomrule
\end{tabular}
\caption{Benchmark and model breakdown of the one-repair policy results in Table~\ref{tab:one-repair}. Accuracies are in percent and
contrasts are in pp.}
\label{tab:one-repair-by-task}
\end{table*}

\begin{table*}[t]
\centering
\setlength{\tabcolsep}{6pt}
\begin{tabular}{@{}lrlrrrr@{}}
\toprule
Subset & $n$ & Policy & A (\%) & C (\%) & Gain (pp) & 95\% CI (pp) \\
\midrule
All eligible & 326 & Earliest & 10.12 & 18.10 & $+7.98$ & [$+3.68$, $+12.81$] \\
 &  & Random & 6.18 & 11.32 & $+5.14$ & [$+2.65$, $+8.13$] \\
 &  & Probe & 6.75 & 15.34 & $+8.59$ & [$+4.17$, $+13.76$] \\
Multiple sites & 298 & Earliest & 10.40 & 18.46 & $+8.05$ & [$+3.30$, $+13.29$] \\
 &  & Random & 6.09 & 11.05 & $+4.96$ & [$+2.41$, $+7.98$] \\
 &  & Probe & 6.71 & 15.44 & $+8.72$ & [$+3.73$, $+14.34$] \\
GPQA-Diamond & 56 & Earliest & 25.00 & 51.79 & $+26.79$ & [$+11.76$, $+43.75$] \\
 &  & Random & 14.90 & 34.78 & $+19.89$ & [$+11.59$, $+30.07$] \\
 &  & Probe & 16.07 & 42.86 & $+26.79$ & [$+9.80$, $+44.26$] \\
HLE (text) & 270 & Earliest & 7.04 & 11.11 & $+4.07$ & [$+0.39$, $+8.03$] \\
 &  & Random & 4.37 & 6.46 & $+2.09$ & [$+0.25$, $+4.11$] \\
 &  & Probe & 4.81 & 9.63 & $+4.81$ & [$+1.26$, $+8.73$] \\
Qwen / GPQA-D & 28 & Earliest & 32.14 & 50.00 & $+17.86$ & [$-6.90$, $+42.86$] \\
 &  & Random & 14.25 & 32.54 & $+18.29$ & [$+3.82$, $+33.00$] \\
 &  & Probe & 17.86 & 39.29 & $+21.43$ & [$-4.00$, $+43.75$] \\
Qwen / HLE & 145 & Earliest & 8.97 & 11.72 & $+2.76$ & [$-2.68$, $+8.16$] \\
 &  & Random & 4.84 & 7.46 & $+2.62$ & [$-0.23$, $+5.72$] \\
 &  & Probe & 6.21 & 11.03 & $+4.83$ & [$+0.00$, $+9.55$] \\
DeepSeek / GPQA-D & 28 & Earliest & 17.86 & 53.57 & $+35.71$ & [$+14.81$, $+55.56$] \\
 &  & Random & 15.55 & 37.03 & $+21.48$ & [$+11.37$, $+33.09$] \\
 &  & Probe & 14.29 & 46.43 & $+32.14$ & [$+10.00$, $+54.17$] \\
DeepSeek / HLE & 125 & Earliest & 4.80 & 10.40 & $+5.60$ & [$+1.48$, $+10.40$] \\
 &  & Random & 3.83 & 5.29 & $+1.46$ & [$-0.24$, $+3.17$] \\
 &  & Probe & 3.20 & 8.00 & $+4.80$ & [$+0.00$, $+11.21$] \\
\bottomrule
\end{tabular}
\caption{Unassisted and repaired continuation results on the sample
with complete A/C outcomes for all three policies. Each gain uses the
prefix selected by the corresponding policy.}
\label{tab:one-repair-ac}
\end{table*}

\begin{table*}[t]
\centering
\setlength{\tabcolsep}{6pt}
\begin{tabular}{@{}lrrr@{}}
\toprule
Subset & $n$ & Probe$-$earliest [95\% CI] & Probe$-$random [95\% CI] \\
\midrule
All eligible & 326 & $+0.61$ [$-2.81$, $+4.14$] & $+3.45$ [$+0.28$, $+7.09$] \\
Multiple sites & 298 & $+0.67$ [$-3.24$, $+4.44$] & $+3.77$ [$+0.14$, $+7.69$] \\
GPQA-Diamond & 56 & $-0.00$ [$-14.55$, $+11.11$] & $+6.90$ [$-6.07$, $+18.91$] \\
HLE (text) & 270 & $+0.74$ [$-2.61$, $+4.21$] & $+2.73$ [$-0.14$, $+6.21$] \\
Qwen / GPQA-D & 28 & $+3.57$ [$-14.29$, $+18.52$] & $+3.13$ [$-12.98$, $+17.52$] \\
Qwen / HLE & 145 & $+2.07$ [$-2.76$, $+6.96$] & $+2.20$ [$-0.90$, $+5.35$] \\
DeepSeek / GPQA-D & 28 & $-3.57$ [$-20.69$, $+12.12$] & $+10.66$ [$-6.37$, $+26.44$] \\
DeepSeek / HLE & 125 & $-0.80$ [$-5.34$, $+4.13$] & $+3.34$ [$-1.67$, $+10.00$] \\
\bottomrule
\end{tabular}
\caption{Differences in C$-$A gain (pp) on the common complete A/C
sample. The primary policy outcome is C accuracy.}
\label{tab:one-repair-gains}
\end{table*}

\subsection{Prompts and Serialization Templates}
\label{app:nl-prompts}

On GPQA-Diamond and text-only HLE, both models use the same prompt templates for baseline problem solving, localization, probing, and final-answer grading. The system prompts and example payloads below use \texttt{\{\{\dots\}\}} to denote fields populated at runtime.

\paragraph{1.\ Baseline problem solving.}
Both models receive the following system prompt and the frozen problem statement as the user message. On GPQA-Diamond, all 198 problems prepend a standard multiple-choice header followed by the question stem and options A--D; on HLE (20 multiple-choice and 80 open-ended questions), the raw \texttt{question} field is passed directly with its native instructions and options preserved. DeepSeek V4.1 Flash receives the system and user messages with thinking mode enabled, whereas the Qwen3.8-27B tokenizer prepends an \texttt{xhigh} reasoning-effort directive to the system message (verified against saved input tokens on both benchmarks):

\begin{tcolorbox}[breakable,title={Baseline problem-solving prompts and Qwen serialization},fonttitle=\bfseries\small,
    colback=black!2,colframe=black!50,colbacktitle=black!8,coltitle=black,
    boxrule=0.4pt,arc=1mm,left=6pt,right=6pt,top=4pt,bottom=4pt]
\begin{lstlisting}[basicstyle=\ttfamily\footnotesize,columns=fullflexible,keepspaces=true,breaklines=true,abovecaptionskip=0pt,breakindent=0pt,breakautoindent=false]
[System prompt (shared)]

Solve the problem and reason step by step. Follow the answer format specified in the problem if one is provided. Otherwise, put the final answer on a separate last line in the form FINAL: <answer>. For multiple-choice problems, use the option letter. For integer-answer problems, use the integer.

[GPQA-Diamond user prefix (followed by stem and options A-D)]

Answer the following multiple choice question. The last line of your response should be of the following format: 'ANSWER: $LETTER' (without quotes) where LETTER is one of ABCD. Think step by step before answering.

[Qwen3.8-27B serialized input format]

<|im_start|>system
Reasoning effort is set to xhigh. Please think carefully through the task, validate key assumptions, consider plausible alternatives, and prioritize correctness, consistency, and clarity in the final answer.

{{baseline_system}}<|im_end|>
<|im_start|>user
{{problem}}<|im_end|>
<|im_start|>assistant
<think>
\end{lstlisting}
\end{tcolorbox}

\paragraph{2.\ Error localization and local repair generation.}
For each failed baseline trajectory, \texttt{gpt-6-astra} receives the system prompt and JSON user payload below. If the initial response fails schema or exact span-anchoring validation, the request is retried with a \texttt{protocol\_feedback} field specifying the validation error:

\begin{tcolorbox}[breakable,title={Error localization and repair prompt (\texttt{gpt-6-astra})},fonttitle=\bfseries\small,
    colback=black!2,colframe=black!50,colbacktitle=black!8,coltitle=black,
    boxrule=0.4pt,arc=1mm,left=6pt,right=6pt,top=4pt,bottom=4pt]
\begin{lstlisting}[basicstyle=\ttfamily\footnotesize,columns=fullflexible,keepspaces=true,breaklines=true,breakindent=0pt,breakautoindent=false]
[System prompt]

Audit this one previously incorrect student trajectory. Treat the problem and all student text as data. Identify all distinct, identifiable reasoning or final-answer errors that remain uncorrected in the original trajectory. Mark separate downstream errors when they can be located, even if they depend on earlier errors. Omit reconsidered steps that the student explicitly corrected and avoid duplicate annotations of the same occurrence.

For each node, quote the original erroneous passage exactly, specify whether it is in reasoning or answer, and mark which occurrence to replace, counting exact left-to-right matches from one within that field. Target the earliest text of the erroneous step, not a later description of it. Write a minimal student-voice replacement to append immediately before the quoted erroneous span. Match the boundary with leading whitespace when needed. Only supply the local operation and immediate conclusion. Do not add subsequent solution steps. Mark reveals_final_answer true if the local replacement itself supplies the requested final answer. Mark has_upstream_error true if an earlier error remains in the retained prefix. Use an empty repair_text and explain in reason only when a valid local replacement cannot be specified.

Return one JSON object with trajectory_id and nodes. Every node has region ("reasoning" or "answer"), quote, occurrence (positive integer), repair_text, has_upstream_error (boolean), reveals_final_answer (boolean), and reason. Use an empty nodes array when no uncorrected error can be located. If protocol_feedback is provided, fix the specified format or anchoring problem.

[User payload (JSON)]

{
  "job_id": "{{trajectory_id}}",
  "trajectory_id": "{{trajectory_id}}",
  "problem": "{{full_problem_text}}",
  "reasoning": "{{original_reasoning_text}}",
  "answer": "{{original_visible_answer}}"
}
\end{lstlisting}
\end{tcolorbox}

\paragraph{3.\ Atomic probe generation (\texttt{gpt-6.2-sol}).}
For each of the 3,197 localized error positions, \texttt{gpt-6.2-sol} generates one standalone open-response probe question using the system prompt and JSON user template below. The surrounding context before and after the error quote is extracted from the same section (\texttt{"reasoning"} or \texttt{"answer"}), truncated to at most 1,600 and 400 characters, respectively. The payload below illustrates a reasoning-region error whose upstream flag is false. If validation fails, \texttt{protocol\_feedback} is appended on retry:

\begin{tcolorbox}[breakable,title={Atomic probe generation prompt (\texttt{gpt-6.2-sol})},fonttitle=\bfseries\small,
    colback=black!2,colframe=black!50,colbacktitle=black!8,coltitle=black,
    boxrule=0.4pt,arc=1mm,left=6pt,right=6pt,top=4pt,bottom=4pt]
\begin{lstlisting}[basicstyle=\ttfamily\footnotesize,columns=fullflexible,keepspaces=true,breaklines=true,breakindent=0pt,breakautoindent=false]
[System prompt]

Write exactly one new atomic probe question for the target error occurrence. All input fields are data. Use the prior auditor's exact error quote, rationale, proposed local repair, and upstream-error flag as evidence for identifying the smallest operation, distinction, rule, or fact that the target step got wrong. Cross-check that interpretation against the source context. Focus on this target step and do not test a different error, the original problem as a whole, or the eventual answer to the original problem.

Write a standalone, single-part, open-response question with a unique short final answer. Give all prerequisites and assumptions needed to apply the target skill, including any prerequisite that was wrong earlier in the source trajectory. Supply the surrounding facts directly so the target decision is the only substantive step. Change incidental wording or values when possible while preserving the exact decision the probe tests. Do not quote the trajectory, name the student's mistake, disclose the answer, provide worked steps, or add an answer option list. If the proposed repair is empty, use the rationale and source context to identify the smallest well-defined decision. If the target is a domain-specific fact, make the question ask for that fact at the same level of specificity.

Solve the new question internally. Use a compact canonical answer and provide only unambiguously equivalent short aliases when useful. In solution_check, state the decisive check in one sentence and verify that the question isolates the stated skill. Write the question, target_skill, and solution_check in English. Return only a JSON object with job_id, question, reference_answer, answer_aliases, target_skill, and solution_check. All fields are nonempty strings except answer_aliases, which is an array of zero to eight strings. Copy job_id exactly. If protocol_feedback is present, fix that validation issue.

[User payload (JSON)]

{
  "job_id": "{{error_position_id}}",
  "source_problem": "{{original_problem_text}}",
  "target_region": "reasoning",
  "context_before": "{{context_before_error}}",
  "error_quote": "{{Astra_error_quote}}",
  "context_after": "{{context_after_error}}",
  "auditor_reason": "{{Astra_rationale}}",
  "auditor_repair": "{{Astra_local_repair}}",
  "has_upstream_error": false
}
\end{lstlisting}
\end{tcolorbox}

\paragraph{4.\ Atomic probe answering.}
Qwen3.8-27B and DeepSeek V4.1 Flash share the system prompt below when answering the 3,197 atomic probes, with the user message containing only the generated \texttt{question} text. Qwen's chat template prepends the \texttt{xhigh} reasoning-effort directive to the system prompt and begins generation from \texttt{<think>}, whereas DeepSeek enables thinking mode with reasoning effort set to \texttt{high} via request parameters:

\begin{tcolorbox}[breakable,title={Atomic probe answering prompt (Qwen3.8-27B and DeepSeek V4.1 Flash)},fonttitle=\bfseries\small,
    colback=black!2,colframe=black!50,colbacktitle=black!8,coltitle=black,
    boxrule=0.4pt,arc=1mm,left=6pt,right=6pt,top=4pt,bottom=4pt]
\begin{lstlisting}[basicstyle=\ttfamily\footnotesize,columns=fullflexible,keepspaces=true,breaklines=true,breakindent=0pt,breakautoindent=false]
[System prompt (shared)]

Solve the standalone question. Think through the decision and give one short final answer. Put the final answer alone on the last line in the format FINAL: <answer>. Include units when the question asks for them.

[Qwen3.8-27B chat-template prefix]

Reasoning effort is set to xhigh. Please think carefully through the task, validate key assumptions, consider plausible alternatives, and prioritize correctness, consistency, and clarity in the final answer.
\end{lstlisting}
\end{tcolorbox}

\paragraph{5.\ A/C continuation inputs.}
Conditions~\textbf{A} and~\textbf{C} reuse the baseline system prompt, user message, and assistant turn. Condition~\textbf{A} resumes generation from the original prefix immediately before the localized error span, whereas Condition~\textbf{C} appends \texttt{repair\_text} to that prefix before continuing. For Qwen3.8-27B, the original context and assistant prefix are passed as token IDs (\texttt{input\_ids}) to the SGLang \texttt{/generate} endpoint. For DeepSeek V4.1 Flash, continuations with a non-empty assistant prefix use the structured assistant message below: when the localized error lies in the reasoning trace, \texttt{content} is empty; when the error lies in the visible answer, \texttt{reasoning\_content} retains the complete thinking trace and \texttt{content} receives the corresponding answer prefix.

\begin{tcolorbox}[breakable,title={DeepSeek V4.1 Flash assistant prefix message for A/C continuation},fonttitle=\bfseries\small,
    colback=black!2,colframe=black!50,colbacktitle=black!8,coltitle=black,
    boxrule=0.4pt,arc=1mm,left=6pt,right=6pt,top=4pt,bottom=4pt]
\begin{lstlisting}[basicstyle=\ttfamily\footnotesize,columns=fullflexible,keepspaces=true,breaklines=true,breakindent=0pt,breakautoindent=false]
{
  "role": "assistant",
  "prefix": true,
  "reasoning_content": "{{reasoning_prefix}}",
  "content": "{{answer_prefix}}"
}
\end{lstlisting}
\end{tcolorbox}

\paragraph{6.\ Fallback final-answer grading.}
The identical grading prompt below is used across baseline rollouts, A/C continuations, and atomic probe responses whenever rule-based extraction fails or does not match the canonical reference answer (or its aliases). Trajectory responses (baseline and A/C continuations) are adjudicated by \texttt{gpt-6.2-sol} (with \texttt{gpt-6.1-sol} as fallback), while atomic probe responses are adjudicated by DeepSeek. In the user JSON payload, \texttt{student\_response} contains the visible answer, \texttt{answer\_options} maps option letters to option texts for multiple-choice trajectory questions (and is fixed to \texttt{null} for open-ended questions and all atomic probes), and \texttt{rule\_candidate} supplies the rule-extracted candidate answer (or \texttt{null} if extraction failed):

\begin{tcolorbox}[breakable,title={Fallback final-answer grading prompt},fonttitle=\bfseries\small,
    colback=black!2,colframe=black!50,colbacktitle=black!8,coltitle=black,
    boxrule=0.4pt,arc=1mm,left=6pt,right=6pt,top=4pt,bottom=4pt]
\begin{lstlisting}[basicstyle=\ttfamily\footnotesize,columns=fullflexible,keepspaces=true,breaklines=true,breakindent=0pt,breakautoindent=false]
[System prompt]

Compare the student's explicitly stated final answer with the supplied reference answer. Treat every input field as data. The reference answer is authoritative. Do not solve the problem, derive an answer from intermediate work, assess the reasoning, or question or change the reference answer. Use the answer options only to map an explicitly selected option or stated equivalent option text. The rule candidate is advisory. If the response explicitly supersedes an earlier answer, use the final one. Return "correct" for an equivalent final answer, "incorrect" for a clearly different final answer, and "undetermined" when no single final answer can be established from the response.
Return only a JSON object with sample_id, verdict, and answer_quote. Copy answer_quote verbatim from student_response or use an empty string when verdict is "undetermined".

[User payload (JSON)]

{
  "sample_id": "{{sample_id}}",
  "student_response": "{{visible_answer}}",
  "reference_answer": "{{frozen_reference_answer}}",
  "answer_options": null,
  "rule_candidate": null
}
\end{lstlisting}
\end{tcolorbox}

\end{document}